\documentclass{article}

\usepackage{PRIMEarxiv}

\usepackage[utf8]{inputenc} % allow utf-8 input
\usepackage[T1]{fontenc}    % use 8-bit T1 fonts
\usepackage{hyperref}       % hyperlinks
\usepackage{url}            % simple URL typesetting
\usepackage{booktabs}       % professional-quality tables
\usepackage{amsfonts}       % blackboard math symbols
\usepackage{nicefrac}       % compact symbols for 1/2, etc.
\usepackage{microtype}      % microtypography
\usepackage{lipsum}
\usepackage{fancyhdr}       % header
\usepackage{graphicx}       % graphics
\graphicspath{{media/}}     % organize your images and other figures under media/ folder

\usepackage[title]{appendix}
\usepackage{amsmath}
\usepackage{comment}
\usepackage{enumitem}
\usepackage{algorithm}
\usepackage{algorithmic}
\usepackage{multirow}
\usepackage{siunitx}
\usepackage{subcaption}
\usepackage{longtable}
\usepackage{tabularx}
\usepackage{xcolor}
\usepackage{array}
\usepackage{makecell}
\usepackage{threeparttable}
\makeatletter
\@ifundefined{showeprint}{}{%
}
\makeatother

\makeatletter
\def\nodot.{} 
\makeatother

\newcommand{\magenta}[0]{\textcolor{magenta}}
\usepackage{xstring}
\usepackage{xurl}

\makeatletter
\let\OrigHref\href
\renewcommand{\href}[2]{%
  \begingroup
  \edef\Hy@t{#1}%
  \IfBeginWith{\Hy@t}{https://doi.org/}%
    {\endgroup\OrigHref{#1}{\nolinkurl{#1}}}%
    {\endgroup\OrigHref{#1}{#2}}%
}
\makeatother

\newcolumntype{M}[1]{>{\centering\arraybackslash}m{#1}}
\newcolumntype{T}{|@{}M{0.13\linewidth}@{}|}
\newcolumntype{Z}{@{}M{0.14\linewidth}@{}|}

\title{TopiCLEAR: Adaptive embedding clustering for interpretable topic discovery from short texts
}

\author{
  Aoi Fujita, Taichi Yamamoto, Yuri Nakayama,  \\
Graduate School of Frontier Sciences, The University of Tokyo\\
Chiba, Japan\\
   \And
   Ryota Kobayashi \\
   Graduate School of Frontier Sciences, The University of Tokyo\\
   Chiba, Japan\\
   Mathematics and Informatics Center, The University of Tokyo\\
   Tokyo, Japan\\
   \texttt{r-koba@edu.k.u-tokyo.ac.jp} \\
}

\begin{document}
\maketitle

\begin{abstract}  
Topic discovery is a fundamental technique for text mining that identifies abstract topics within large document collections. 
A recent approach to topic discovery is to cluster document or sentence embeddings, typically obtained from pre-trained language models, and represent each cluster as a topic. Despite their strong empirical performance, the design principles linking the geometry of embedding spaces to human-interpretable topic organization remain unclear. Clarifying the relationship between these geometric structures and topic interpretability is therefore a key challenge in topic discovery. 
In this study, we propose TopiCLEAR (Topic discovery by CLustering Embeddings with Adaptive dimensionality Reduction), a simple framework that integrates document embeddings with iterative clustering based on adaptive dimensionality reduction. 
TopiCLEAR is guided by the hypothesis that human-interpretable topics correspond to low-dimensional geometric structures in embedding spaces and {leverages adaptive dimensionality reduction to identify them.}
We evaluate topic quality using a document-level approach that combines quantitative evaluation based on human-labeled data with qualitative assessment in the absence of ground-truth labels. 
Experiments on four benchmark datasets, covering both formal and informal texts, show that TopiCLEAR consistently achieves strong agreement with human annotations, particularly for short and informal texts. 
Furthermore, a case study on Twitter data demonstrates that TopiCLEAR produces more interpretable topics than Latent Dirichlet Allocation (LDA), recovering both human-annotated topic structure and coherent sub-topic structure. 
These results highlight the effectiveness of {clustering documents in a low-dimensional topic space} for topic discovery.
%% These results highlight the effectiveness of jointly learning low-dimensional topic representations and clustering for topic discovery.
\end{abstract}

% keywords can be removed

\section{Introduction}

%%%%%%%%%%%%%%%%%%%%%%%%%%%%%%
%%%%%  Topic Discovery
%%%%%%%%%%%%%%%%%%%%%%%%%%%%%%
Topic discovery, the task of identifying abstract topics within a large collection of documents, has been extensively studied in text mining and natural language processing~\cite{blei2012probabilistic,zhao2021topic,Qiang_2022}. 
Topic discovery techniques have been successfully applied across diverse domains, including the analysis of scientific journal papers~\cite{blei2012probabilistic,yau2014clustering}, user opinions~\cite{hemmatian2019survey}, public perception of social issues~\cite{kobayashi_evolution_2022,miyazaki-2024}, and political discourse and speeches~\cite{yaqub-2017,walter2021strategy}.

%%%%%%%%%%%%%%%%%%%%%%%%%%%%%%
%%%%%   1st approach: Topic models -- Word level Probabilistic Model
%%%%%%%%%%%%%%%%%%%%%%%%%%%%%%
A standard approach to topic discovery is topic modeling~\cite{blei2012probabilistic}, which are probabilistic models that generate documents word by word. In this approach, topics are inferred from Bag-of-Words (BoW) representations. 
Topic modeling began with Latent Dirichlet Allocation (LDA)~\cite{blei_latent_2003} and has evolved through extensions in word distributions~\cite{blei2012probabilistic} and inference methods~\cite{zhao2021topic}. 
These models have demonstrated success in analyzing well-structured, longer documents such as news articles and academic texts~\cite{blei2012probabilistic,zhao2021topic}. 
However, for short or informal texts~\cite{Qiang_2022}, interpreting topics becomes challenging. 
Topic models represent topics using representative words (e.g., 'top words'), but such word-level representations may fail to capture context, especially with synonyms, lexical variations, or misspellings~\cite{Qiang_2022}.

%%%%%%%%%%%%%%%%%%%%%%%%%%%%%%
%%%%%  2nd approach: Clustering approach
%%%%%%%%%%%%%%%%%%%%%%%%%%%%%%
Another approach discovers topics by directly clustering document or sentence embeddings, typically obtained from pre-trained language models, and then representing each cluster as a topic. 
Methods such as BERTopic and Top2Vec combine embedding spaces with density-based clustering to identify topics~\cite{Maarten_2022,Angelov_2020,yang2025enhancing}. 
Despite their strong empirical performance, the design principles linking the geometry of embedding spaces to human-interpretable topic organization remain unclear. 
Clarifying the relationship between these geometric structures and topic interpretability is therefore a key challenge in topic discovery.

In this study, we {present} a simple framework for topic discovery, TopiCLEAR (Topic discovery by CLustering Embeddings with Adaptive dimensionality Reduction). This method is guided by the hypothesis that human-interpretable topics correspond to low-dimensional geometric structures in embedding spaces. 
First, TopiCLEAR embeds each text using Sentence-BERT (SBERT) and obtains an initial clustering.  
It then {constructs} a low-dimensional topic space---a subspace of the original embedding space---via iterative clustering refinement based on Adaptive Dimensionality Reduction (ADR). \\

In addition, evaluating topic quality remains an important challenge, particularly when dealing with short and informal texts such as social media data. 
To address this, we consider two approaches to evaluating topic quality.
When human labels are available, we measure the alignment between topics labeled by humans and those identified by the method. In the absence of such labels, topic quality is assessed by examining representative documents.

Our contributions are twofold:
\begin{enumerate}   
  \item     We {present TopiCLEAR} a simple topic discovery {framework that identifies topics through clustering in a low-dimensional topic space obtained via adaptive dimensionality reduction.}
  \item     We provide an evaluation approach for topic discovery combining alignment with human-labeled data and qualitative analysis without ground-truth labels.
\end{enumerate}

The rest of the paper is organized as follows. Section 2 reviews existing studies for topic discovery and evaluation. Section 3 presents the proposed method, TopiCLEAR. Section 4 describes the datasets, baselines, and evaluation metrics. Section 5 reports quantitative evaluations and ablation studies of TopiCLEAR, while Section 6 presents qualitative evaluation on a Twitter dataset. Section 7 concludes and discusses future research directions.

\section{Related Work}

This section reviews related work on topic discovery methods and evaluation metrics for topic quality.

\subsection{Topic Discovery}

Topic discovery refers to the automatic identification of underlying themes (or topics) within large volumes of text, such as news articles, academic papers, and books. 
In topic discovery, texts (e.g., news articles or social media posts) are referred to as \textit{documents}; we follow this convention throughout the paper.  
Among the various techniques for topic discovery, statistical approaches known as topic models~\cite{blei2012probabilistic,vayansky_2020} are the most widely used. 
Latent Dirichlet Allocation (LDA)~\cite{blei_latent_2003} is a classical topic model. It models each topic as a distribution over words and each document as a mixture of latent topics, inferred from Bag-of-Words (BoW) representations (i.e., word count data). 
Due to its simplicity, LDA has led to various extensions.
Biterm Topic Model (BTM)~\cite{Yan_2013} exploits corpus-level word co-occurrence patterns to better handle short texts. 
Furthermore, several models have been proposed to incorporate semantic information via text embeddings. 
For example, the embedding-based topic model~\cite{qiang2017topic} discovers topics from aggregated pseudo-documents using word embeddings. 
The Embedded Topic Model (ETM)~\cite{Dieng_2020} leverages the inner product between word embeddings and topic vectors. 
The Contextualized Topic Model (CTM)~\cite{Bianchi_2021} combines BoW representations with contextual embeddings to infer topics more effectively.
In addition, Neural Topic Models (NTMs)~\cite{Srivastava_2017} (see review~\cite{zhao2021topic}) leverage deep neural networks to enhance modeling flexibility and capture complex topic–word relationships. 
However, NTMs may (i) be sensitive to preprocessing choices (e.g., stop‑word removal), (ii) involve variational‑inference training, and (iii) typically require GPU‑level computation. 
Unlike NTMs, our {framework} requires neither heuristic preprocessing nor gradient‑based training and does not require GPUs. 

In contrast to topic models, another line of work identifies topics by clustering document embeddings~\cite{Maarten_2022,Angelov_2020,yang2025enhancing}. 
For example, BERTopic~\cite{Maarten_2022} encodes documents with a pre-trained language model (e.g., SBERT), and employs HDBSCAN for clustering. 
Similarly, Top2Vec~\cite{Angelov_2020} embeds both documents and words in a shared vector space and uses UMAP and HDBSCAN to identify topics. 
Our {framework} follows the embedding-clustering paradigm but differs in its focus on the clustering strategy. 
More specifically, TopiCLEAR improves topic separability by {constructing} a low-dimensional topic space from cluster assignments (via ADR), rather than relying on manual preprocessing or changing the embedding model. 
The method design is motivated by the topic representation hypothesis, which posits that human-interpretable topics are associated with low-dimensional geometric structures in embedding spaces.

We refer to methods that infer topics algorithmically from numerical document representations (e.g., BoW or embeddings) as \emph{representation-based} approaches. 
Recently, \emph{prompt-driven} LLM-based approaches have emerged, which prompt an LLM with instructions (together with a document collection) to produce topic descriptions and/or document--topic assignments~\cite{pham2024topicgpt,liu2025llm}. 
From a benchmarking perspective, these pipelines introduce additional degrees of freedom (e.g., prompt templates~\cite{razavi2025benchmarking} and generation settings) that are not directly analogous to hyperparameter choices in representation-based approaches. 
Accordingly, in this paper we benchmark \emph{representation-based} methods under controlled and reproducible settings, 
and treat \emph{prompt-driven} LLM-based approaches as related work rather than benchmarking baselines.

\subsection{Topic Quality Evaluation}

%%%%%%%%%%%%%%%%%%%%%%%%%%%%%%
%%%%%  Topic coherence 
%%%%%%%%%%%%%%%%%%%%%%%%%%%%%%
A common approach to evaluating topic quality is to use topic coherence metrics, which aim to quantify the semantic coherence based on representative words (i.e., top words). 
%% UCI 
One of the earliest metrics is UCI coherence~\cite{Newman_2010}, defined as the average Pointwise Mutual Information (PMI) of word pairs that frequently occur within a topic. A high UCI score indicates that top words in a topic tend to co-occur in the corpus. 
%%  NPMI
Aletras and Stevenson~\cite{Aletras_2013} extended the UCI metric by introducing Normalized PMI (NPMI), which showed stronger alignment with human judgments. 
%%  Cv   
Furthermore, Röder et al.~\cite{roder_2015} proposed the $C_v$ metric by systematically exploring various coherence metrics across public datasets. 
These metrics have become standard tools for evaluating and tuning topic models. 

However, evaluating topic quality using top words becomes more difficult when dealing with short texts, such as social media posts~\cite{Qiang_2022}.
In such cases, word co-occurrence is sparse, and coherence metrics that rely on co-occurrence statistics may fail to evaluate the interpretability of topics. 
It remains challenging to evaluate the relationship between various coherence metrics and human evaluations, including interpretability. 
For example, Doogan and Buntine~\cite{doogan_2021} found no correlation between coherence metrics and human interpretability in a tweet corpus. 
Similarly, Hoyle et al.~\cite{Hoyle_2021} observed that even when coherence scores indicated significant differences between models, human annotators were unable to perceive those differences, questioning the validity of coherence-based comparisons. 
In contrast, Lim et al.~\cite{Lim_2024} demonstrated that tuning the hyperparameters of coherence metrics can substantially improve their correlation with human evaluations.

In machine learning, topic discovery from text is typically treated as an unsupervised learning task, specifically clustering. 
Since ground-truth labels for each document are unavailable, evaluating the quality of discovered topics remains a challenging problem. 
Nevertheless, when reliable clustering results are available, such as those obtained through human annotation, metrics such as the Adjusted Mutual Information (AMI)~\cite{vinh_information_2009} and the Adjusted Rand Index (ARI)~\cite{Hubert_1985} can be used to assess clustering performance.

\section{Proposed {Framework}: TopiCLEAR}
We {present} TopiCLEAR, a {framework} for discovering $K$ topics from $N$ documents based on their textual content, where the number of topics $K$ is assumed to be given. 

{Our framework} consists of three steps: 
\begin{itemize}
    \item   Compute document embeddings.
    \item   Obtain an initial clustering.
    \item   Iteratively refine the clustering.    
\end{itemize}
TopiCLEAR first calculates embeddings for each document. 
It then obtains an initial clustering by applying principal component analysis (PCA) for dimensionality reduction and 
{apply an clustering algorithm.} 
Finally, the initial clustering is iteratively refined using Adaptive Dimensionality Reduction (ADR) to improve cluster separability.  
{For concreteness, we describe TopiCLEAR using Gaussian Mixture Model (GMM) as the clustering algorithm. Hereinafter, we will refer to this implementation as TopiCLEAR(GMM). Algorithm~\ref{alg:adr} provides its pseudo-code, and the following subsections describe each step in detail.
The proposed framework is generic and can be readily extended to other clustering methods, including $k$-means.}   
%%  Algorithm~\ref{alg:adr} provides its pseudo-code, and the following subsections describe each step in detail. 
\begin{algorithm}[tb]
  \caption{TopiCLEAR(GMM) for discovering topics from text}
  \label{alg:adr}
  \begin{algorithmic}[1]
    \REQUIRE document \textit{text}, number of topics $K$, feature dimension $D$
    \STATE $X_{\rm raw} \leftarrow \mathrm{SBERT}(\textit{text})$
    \STATE $X_D \leftarrow \mathrm{PCA}(X_{\rm raw}, D)$
    \STATE $Y_D \leftarrow \mathrm{normalize}(X_D)$
    \STATE $Y_{K-1} \leftarrow \mathrm{PCA}(Y_D, K-1)$
    \STATE $H \leftarrow \mathrm{GMM}(Y_{K-1})$
    \REPEAT
      \STATE $\hat{H} \leftarrow H$
      \STATE $Y_{K-1} \leftarrow \mathrm{FDA}(Y_D, H)$
      \STATE $H \leftarrow \mathrm{GMM}(Y_{K-1})$
    \UNTIL{$\hat{H}=H$}
    \STATE \textbf{return} $H$
  \end{algorithmic}
\end{algorithm}

\subsection{Document Embeddings}
\label{section_embeddings}

We use Sentence-BERT (SBERT)~\cite{reimers_sentence-bert_2019} to embed each document into a fixed-size vector (Algorithm~\ref{alg:adr}, line~1). 
Unlike traditional word-level embeddings, SBERT encodes entire documents into single fixed-size vectors, enabling all documents to be embedded in a shared semantic space. 
We adopt the \texttt{all-MiniLM-L6-v2}\footnote{\url{https://huggingface.co/sentence-transformers/all-MiniLM-L6-v2}} model from Hugging Face to generate 384-dimensional embeddings.  
This model is based on a pre-trained MiniLM architecture~\cite{Wang_2020} and has been fine-tuned on over one billion sentence pairs from diverse datasets. 
Let $\mathbf{x}_{{\rm raw},n}$ be the embedding of document $n$, and $X_{\rm raw}$ be the matrix of all embeddings: $X_{\rm raw}=[\mathbf{x}_{{\rm raw},1},\dots,\mathbf{x}_{{\rm raw},N}]$.

\subsection{Initial Clustering}

We obtain an initial clustering result from the embeddings.  
First, we apply PCA on the embeddings $\mathbf{x}_{{\rm raw},n}$ to obtain $D$-dimensional vectors $\mathbf{x}_{D,n}$ and then normalize the vectors to obtain $\mathbf{y}_{D,n}$ (Algorithm~\ref{alg:adr}, lines 2--3): 
\begin{align}
    \mathbf{x}_{D,n} &= \mathbf{W}_{\rm PCA}^{\top}\mathbf{x}_{{\rm raw},n},\\
    \mathbf{y}_{D,n} &= \mathbf{x}_{D,n}/\|\mathbf{x}_{D,n}\|,
\end{align}
where $\mathbf{W}_{\rm PCA}$ is a matrix formed by stacking the top $D$ principal vectors obtained from $X_{\rm raw}$ and $\|\mathbf{x}_{D,n}\|$ denotes the $L^2$ norm of the vector $\mathbf{x}_{D,n}$. 
{The feature dimension $D$ is set to $D=64$, which was selected as a practical trade-off between accuracy and computational efficiency (see Table~\ref{tab:ablation_pre_ami}).}
%% The PCA dimension $D$ has only a minor influence on the resulting topic quality (Section~\ref{section_ablation}). 
Similar to the embeddings, we obtain the data matrix by concatenation: $X_D=[\mathbf{x}_{D,1},\dots,\mathbf{x}_{D,N}]$ and $Y_D=[\mathbf{y}_{D,1},\dots,\mathbf{y}_{D,N}]$.

The normalized features $Y_D$ are further reduced to $(K-1)$ dimensions, yielding $Y_{K-1}=[\mathbf{y}_{K-1,1},\dots,\mathbf{y}_{K-1,N}]$, to obtain a compact topic space (Algorithm~\ref{alg:adr}, lines 4--5). 
We then cluster the reduced features $Y_{K-1}$ using a Gaussian Mixture Model (GMM) that assumes each topic follows a Gaussian distribution:
\begin{align*}
  P[z_n=k] &= \pi_k, \\
  p(\mathbf{y}_{K-1,n} | z_n=k)
  &\propto    
    \exp\left(
        -\tfrac{1}{2}
        (\mathbf{y}_{K-1,n}-\boldsymbol{\mu}_k)^\top
        \boldsymbol{\Sigma}_k^{-1}
        (\mathbf{y}_{K-1,n}-\boldsymbol{\mu}_k)
    \right), 
\end{align*}
where $z_n$ is a latent variable representing the topic of the document $n$ and $\pi_k$, $\boldsymbol{\mu}_k$, and $\boldsymbol{\Sigma}_k$ are the mixture weights, mean, and covariance for a topic $k=1,\dots,K$, respectively. 
%  GMM: Parameter Fitting
The parameters $\{\pi_k,\boldsymbol{\mu}_k,\boldsymbol{\Sigma}_k\}$ are fitted by maximizing the likelihood using the EM algorithm, assuming diagonal covariance matrices for $\boldsymbol{\Sigma}_k$. The posteriors $\gamma_{n,k}=P[z_n=k \mid \mathbf{y}_{K-1,n}]$ are then calculated based on the fitted parameters. 
%  Initial Topic assignment
Finally, we assign each document to the maximum-posterior topic,
$h_n=\operatorname{argmax}_{k}\gamma_{n,k}$, and obtain the topic vector $H=\{h_n\}_{n=1}^N$. 
We use the implementation in the Python package
\texttt{scikit-learn}~\cite{pedregosa_scikit-learn_2011} for these computations.

\subsection{Iterative Refinement} 

We refine the initial features $Y_{K-1}$ and the topic assignments $H$ (Algorithm~\ref{alg:adr}, lines 6--10). 
Our approach is based on \emph{Adaptive Dimensionality Reduction} (ADR)~\cite{Ding_2007}, which alternates between (A) extracting a low-dimensional feature space $Y_{K-1}$ using the current clustering result and (B) clustering using the updated features $Y_{K-1}$. 
Intuitively, ADR treats the current cluster assignments $H$ as pseudo-labels to learn a projection that increases topic separability.

In step A), Fisher Discriminant Analysis (FDA; i.e., linear discriminant analysis)~\cite{Fukunaga_1990} is employed for extracting the features $Y_{K-1}$ from the normalized embeddings $Y_D$ and the current assignments $H$. The feature dimension is set to $K-1$, because FDA with $K$ topic labels can extract at most $K-1$ dimensional features~\cite{Fukunaga_1990}. 
FDA performs linear feature extraction by computing $Y_{K-1} = \mathbf{W}_{\rm FDA}^{\top}\,Y_D$, where $\mathbf{W}_{\rm FDA}$ is a transformation matrix optimized to maximize topic separability in the projected space. The objective function $J(\mathbf{W})$ to be maximized is defined as: 
\begin{equation}
    J(\mathbf{W}) =
    \frac{\operatorname{tr}(\mathbf{W}^{\top}\mathbf{S}_B\mathbf{W})}
    {\operatorname{tr}(\mathbf{W}^{\top}\mathbf{S}_W\mathbf{W})},  
    \label{eq:FDA_trace_ratio}
\end{equation}
Here, $\mathbf{S}_W$ and $\mathbf{S}_B$ denote the within-class and between-class scatter matrices, respectively, and are defined as: 
\begin{eqnarray}		
	\mathbf{S}_W &=& \sum_{k=1}^{K}\sum_{n \in C_k}
         (\mathbf{y}_{D, n} -\bar{\mathbf{y}}_{D, k} )(\mathbf{y}_{D, n} -\bar{\mathbf{y}}_{D, k})^{\top},  \\
	  \mathbf{S}_B &=& \sum_{k=1}^{K} N_k (\bar{\mathbf{y}}_{D, k}-\bar{\mathbf{y}}_{D} )(\bar{\mathbf{y}}_{D, k}-\bar{\mathbf{y}}_{D} )^{\top},  
\end{eqnarray} 
where $C_k=\{n:\,h_n=k\}$ is the set of documents assigned to topic $k$, 
$N_k = |C_k|$ is the number of documents in topic $k$, $\bar{\mathbf{y}}_{D} = \frac{1}{N}\sum_{n=1}^{N}\mathbf{y}_{D,n}$ is the global mean, and $\bar{\mathbf{y}}_{D, k} = \tfrac{1}{N_k}\sum_{n \in C_k} \mathbf{y}_{D, n}$ is the class mean of topic $k$. 
In step B), we recluster the updated features $Y_{K-1}$ with the same GMM procedure to obtain a new assignment $H$.

ADR iterates steps A) and B) until the resulting assignment $H$ no longer changes. 
Ding and Li~\cite{Ding_2007} report that their ADR variant (they use $k$-means for clustering, instead of GMM) typically converges in about 10 iterations. Accordingly, we set the maximum number of iterations to 10 for all experiments.  
We adopt ADR with GMM for clustering, as it yields higher topic quality in our ablation results (see Section~\ref{section_ablation} for details). 
\vspace{0.5cm}

\section{Experimental Setup}
This section describes the datasets used in this study, the baseline methods employed for comparison with our {framework} (TopiCLEAR), and the metrics used to evaluate the quality of topic discovery.

\subsection{Datasets}
\label{section_datasets}
We evaluated the performance of TopiCLEAR using the following four datasets. 
These datasets were selected to include both formal texts, such as online news articles, and informal texts, such as social media posts. 
The variety of datasets allows us to evaluate the applicability of TopiCLEAR across varying linguistic styles and document lengths. 
All datasets include ground-truth topic categories. 
The topic categories in 20News, AgNewsTitle, and Reddit are derived from user- or publisher-defined sources. 
In contrast, the categories in TweetTopic are obtained through manual annotation.

Table~\ref{tab:datasets} summarizes the statistical information for these datasets.  
Both 20News and AgNewsTitle are composed of data from online news articles. 
The average document length of 20News (135 words) is much larger than that of AgNewsTitle (5 words). 
Reddit and TweetTopic consist of social media posts that contain a lot of slang and colloquial expressions. 
A brief description of these datasets is provided below: 

\begin{table}[t]
\centering
\caption{Statistics of four datasets used in this study}
\label{tab:datasets} 
\begin{tabularx}{\columnwidth}{X c c c}
\toprule
\textbf{Dataset} &
{\makecell{\textbf{Number of}\\\textbf{Topics}}} &
{\makecell{\textbf{Average}\\\textbf{Words/Doc.}}} &
{\makecell{\textbf{Vocabulary}\\\textbf{Size}}} \\
\midrule
20News & 20 & 135.2 & 36,037 \\
AgNewsTitle & 4 & 5.2 & 11,513 \\
Reddit & 5 & 34.6 & 15,945 \\
TweetTopic & 6 & 12.3 & 3,100 \\
\bottomrule
\end{tabularx}
\end{table}

\textbf{20News~\cite{Lang_1995}:} $\quad$
    It contains 18,806 online news articles. 
    The topic categories are based on the newsgroup names posted to Usenet. 
    Each article has been assigned to one of 20 topic categories, including [medicine], [gun issues], and [Middle East politics].     
    
\textbf{AgNewsTitle~\cite{zhang_character-level_2015}:} $\quad$
    It contains 119,794 online news article titles. 
    Each article is assigned to one of four topic categories: [World], [Science/Technology], [Sports], and [Business]. 
    While the original dataset also includes the texts of articles, this study focuses exclusively on the titles to assess performance on short texts.     
    
\textbf{Reddit~\cite{Curiskis_2019}:} $\quad$
    It consists of parent comments and their replies (total 37,933 posts) from Reddit subreddit pages. 
    The topic categories are based on the names of the subreddit page posted by the users. 
    Each comment is assigned to one of the following five categories based on the subreddit page: [PC], [News], [Movies], [American Football], and [Human Relationships].      
    
\textbf{TweetTopic~\cite{antypas_2022}:} $\quad$ 
    It contains 6,997 Twitter posts. 
    To obtain the topic categories, a manual annotation on Amazon Mechanical Turk was conducted by five annotators and topics that were agreed upon by at least two annotators were retained. The categories were then merged into six non-overlapping topics, and tweets assigned to multiple categories were excluded.     
    Finally, the posts were categorized into one of the following six topics: [art \& culture], [business], [pop culture], [daily life], [sports \& gaming], and [science \& technology].

\subsection{Baseline methods for comparison}
\label{section_baselines}
This study compared TopiCLEAR with five existing methods as baselines. The conventional methods are classified into 
two groups: (i) topic models and (ii) clustering approaches using document or sentence embeddings.

\subsubsection{Topic models}
Topic models are a statistical tool that identfies the ``topics'' within a set of documents. We chose five topic models as baselines. 

First, we adopted Latent Dirichlet Allocation (LDA)~\cite{blei_latent_2003}, a standard topic model based on word distribution. 
Next, we adopted Biterm Topic Model (BTM)~\cite{Yan_2013}, which incorporates word pair distributions for handling short texts, as well as ProdLDA~\cite{Srivastava_2017}, which was modified to allow for a mixture of topics within individual words. 
Furthermore, we included the Embedded Topic Model (ETM)~\cite{Dieng_2020} and the Contextualized Topic Model (CTM)~\cite{Bianchi_2021}, which incorporate both word distributions and embeddings. 
In this study, the word embeddings used in ETM were learned from the dataset. As with TopiCLEAR, we use SBERT for the document embeddings used in CTM.  

%%  Preprocessing
Three preprocessing procedures were performed for these methods: \\
1) Removal of stopwords (e.g., ``the'', ``is'', ``and'', etc.). We used \\ \texttt{nltk.stopwords.words}~\footnote{\url{https://github.com/nltk/nltk}} and 
\texttt{stop\_words.get\_stop\_words}~\footnote{\url{https://github.com/Alir3z4/stop-words}}. \\
2) Removal of symbols and special characters (e.g., ``=='' and ``'s''). \\
3) Exclusion of words occurring five times or less, and words composed solely of non-alphanumeric characters. \\
We used the implementation provided by OCTIS~\cite{Terragni_2021} for LDA, ProdLDA, ETM, and CTM, and the implementation provided by the original authors for BTM\footnote{\url{https://github.com/markoarnauto/biterm}}.

\subsubsection{Language model-based methods}

We adopted BERTopic~\cite{Maarten_2022} as a reference method based on a language model, similar to TopiCLEAR. 
BERTopic discovers topics by clustering document embeddings using the HDBSCAN algorithm~\cite{Campello_2015}. 
We used the original implementation~\footnote{\url{https://github.com/MaartenGr/BERTopic}} and SBERT was used to generate document embeddings for both TopiCLEAR and BERTopic. 
Note that BERTopic automatically determines the number of topics $K$ based on the density of the embedding space. Since other methods, including TopiCLEAR, require $K$ to be specified in advance, the comparison is not strictly controlled, and results should be interpreted with this difference in mind. 

As noted in the Related Work, prompt-based LLM approaches to topic discovery~\cite{pham2024topicgpt,liu2025llm} have recently gained increasing attention. However, a fair comparison with our {framework} is challenging because such approaches use text as both input and output. Therefore, we exclude these approaches from the baselines to ensure a fair and reproducible comparison. 
% \clearpage

\subsection{Evaluation metrics for discovered topics}
\label{section_measures}

We evaluate topic quality using two metrics that measure the similarity between clustering results: Adjusted Mutual Information and Adjusted Rand Index.

\textbf{Adjusted Mutual Information (AMI)}~\cite{vinh_information_2009}: $\ $ 
The AMI score is based on the mutual information, correcting for chance agreement between the two partitions. 
AMI is defined as follows: 
\[
\text{AMI} = \frac{MI(U, V) - \mathbb{E}[MI(U, V)]}{ \bar{H}(U, V) - \mathbb{E}[MI(U, V)] },
\]
where $U= \{ U_1, U_2, \cdots, U_{K_U} \}$ and $V= \{ V_1, V_2, \cdots, V_{K_V} \}$ ($U_i$ and $V_i$ denote the sets of documents assigned to topic $i$) represent the clustering results, $MI(U, V)$ is the mutual information between the two partitions, and $\mathbb{E}[MI(U, V)]$ is its expectation. 
$\bar{H}(U, V):= (H(U)+H(V))/2$ denotes the average of the entropies, where $H(U)$ and $H(V)$ are the entropies of the two clusterings.  
The AMI score is 1 when two clustering results are identical and 0 when they are as similar as if clustered randomly. When the similarity is less than that expected by chance, AMI is negative.

\textbf{Adjusted Rand Index (ARI)}\cite{Hubert_1985}: $\ $ 
The ARI score is based on the Rand Index (RI), correcting for chance agreement between the two partitions. 
ARI is defined as follows: 
\[
\text{ARI} = \frac{RI - \mathbb{E}[RI]}{1 - \mathbb{E}[RI]},
\]
where $RI$ is the Rand Index and $\mathbb{E}[RI]$ is its expectation. 
RI measures the proportion of decisions that are consistent between two clusterings: $\text{RI}= (a+b)/ n_{pair}$, where $a$ $(b)$ represents the number of pairs assigned  to the same (different) clusters in both clusterings, respectively, $n_{pair}= N(N-1)/2$ is the number of pairs, and $N$ is the number of documents in a dataset. 
As with the AMI, the ARI score is 1 when two clustering results are identical and 0 when they are as similar as if clustered randomly. When the similarity is less than that expected by chance, ARI is negative. 
\clearpage

\section{Quantitative Evaluation of TopiCLEAR} 

We quantitatively evaluate topic discovery methods by measuring the degree of agreement with human annotations, which are regarded as a proxy for interpretability. 
In Section~5.1, {our framework, TopiCLEAR, is compared} with five baseline methods on four datasets covering both formal texts, such as news articles, and informal texts, such as social media posts. 
Furthermore, Section~5.2 presents an ablation study of TopiCLEAR to analyze the contributions of the embedding and clustering components.
% \vspace{0.5cm}

\subsection{Comparison with baseline methods} 
\label{section_quantitative}

We compare the performance of {our framework} against five baseline methods and a reference method (Section~\ref{section_baselines}) using two clustering metrics (Section~\ref{section_measures}) across four datasets (Section~\ref{section_datasets}). 
Here, we do not use coherence metrics to evaluate topic quality, as they may not be robust to label noise (Appendix~\ref{appendix_coherence}). 
Table~\ref{tab:score_AMI} presents the AMI scores for all methods. 
TopiCLEAR(GMM) achieves consistently higher AMI and ARI scores than the baseline methods (Appendix~\ref{appendix_ari} for ARI results). 
These results suggest that 
{TopiCLEAR} produces topic clusters that closely match human topic classifications, even for short or informal texts such as news titles and social media posts. 
Note that the comparison with BERTopic is not fully controlled. While BERTopic automatically determines the number of topics $K$ {via HDBSCAN}, the other methods, including TopiCLEAR, assume a fixed value of $K$. 
{
Further analysis revealed substantial discrepancies between the topic granularities inferred by BERTopic and those defined by the human annotations (Appendix~\ref{appendix_ari}). This finding highlights the difficulty of estimating human-interpretable topic granularity from document embeddings. 
Therefore, our quantitative evaluation focuses on agreement with the human annotations under the topic granularity defined by those annotations.  
Here, BERTopic is treated as a reference method operating under a different assumption regarding topic-number determination.
}

Below, we briefly summarize the performance of the baseline methods. 
LDA performed relatively well on the 20News dataset (ranking 4th), but its performance deteriorated dramatically on shorter texts (AgNewsTitle) and social-media datasets (Reddit and TweetTopic). 
BTM, which incorporates word-pair distributions to better handle short texts, indeed achieved stronger results on short documents, but its performance remained limited on colloquial datasets (Reddit and TweetTopic: $\text{AMI} < 0.3$). 
%% While BERTopic performed better than the five baselines except for a short-text dataset (AgNewsTitle), it tends to over-segment topics, which leads to lower ARI scores (Appendix~\ref{appendix_ari}). 
% \newpage

\begin{table}[t]
\centering
\begin{threeparttable}
\caption{{Topic discovery quality quantified by Adjusted Mutual Information (AMI). 
Results are reported as mean $\pm$ standard deviation over five runs. The highest mean value is shown in {\bf bold}. }
}
\label{tab:score_AMI}
\vspace{0.3cm}
\begingroup
\footnotesize
\begin{tabular}{@{}lcccc@{}}
\toprule
\textbf{Method} & \textbf{20News} & \textbf{AgNewsTitle} & \textbf{Reddit} & \textbf{TweetTopic} \\ 
\midrule
LDA & 0.378 $\pm$ 0.014 & 0.042 $\pm$ 0.011 & 0.232 $\pm$ 0.035 & 0.037 $\pm$ 0.018 \\
BTM & 0.419 $\pm$ 0.020 & 0.372 $\pm$ 0.036 & 0.284 $\pm$ 0.012 & 0.181 $\pm$ 0.019 \\
ProdLDA & 0.285 $\pm$ 0.012 & 0.142 $\pm$ 0.018 & 0.093 $\pm$ 0.009 & 0.158 $\pm$ 0.010 \\
ETM & 0.462 $\pm$ 0.022 & 0.027 $\pm$ 0.015 & 0.244 $\pm$ 0.103 & 0.213 $\pm$ 0.015 \\
CTM & 0.315 $\pm$ 0.009 & 0.219 $\pm$ 0.008 & 0.131 $\pm$ 0.010 & 0.226 $\pm$ 0.008 \\
BERTopic\tnote{a} & 0.404 $\pm$ 0.006 & 0.201 $\pm$ 0.002 & 0.268 $\pm$ 0.001 & 0.234 $\pm$ 0.004 \\
TopiCLEAR(GMM) & \textbf{0.588 $\pm$ 0.006} & \textbf{0.489 $\pm$ 0.000} & \textbf{0.462 $\pm$ 0.000} & \textbf{0.382 $\pm$ 0.004} \\
\bottomrule

\end{tabular}
\endgroup
\begin{tablenotes}[flushleft]

\footnotesize
\item[a]  BERTopic is not directly comparable to the other methods, because it does not allow us to specify the number of topics beforehand.    
\end{tablenotes}
\end{threeparttable}
\end{table}

\begin{figure*}[t]
  \centering
  \includegraphics[width= 15cm]{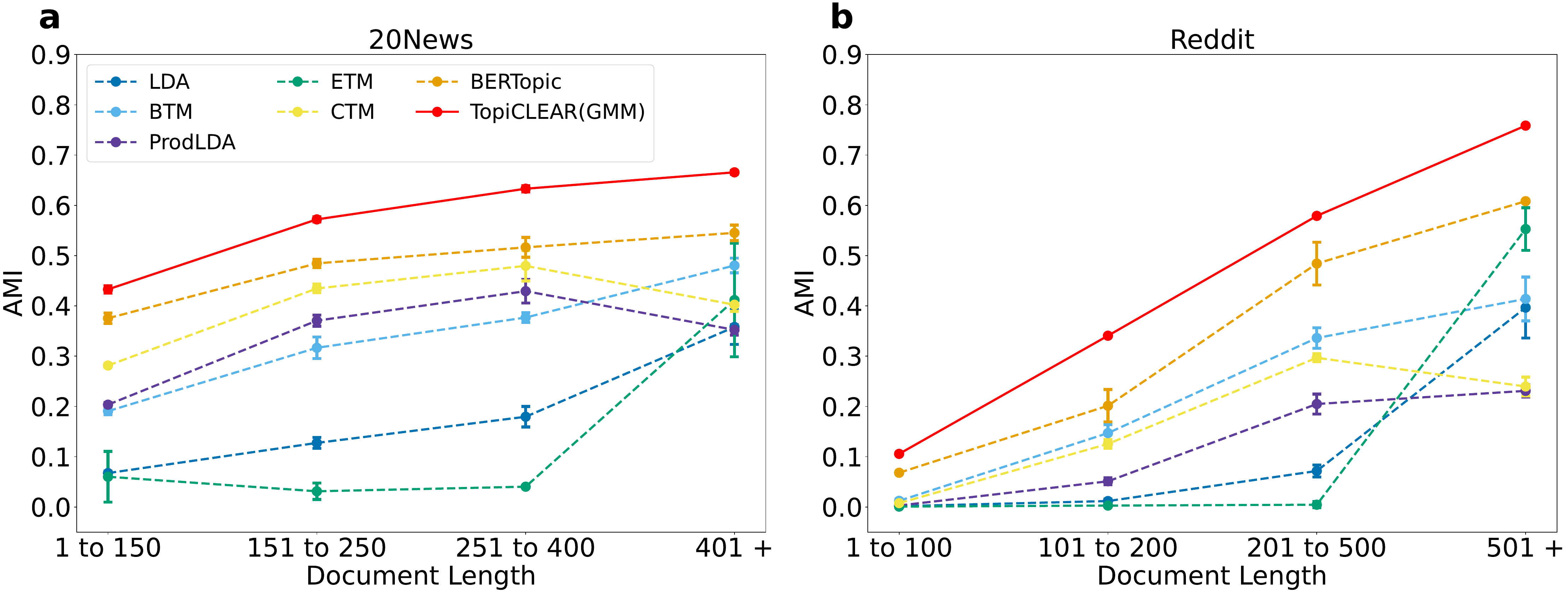}
  \caption{
  Dependence of the AMI score on document length (word count). Two datasets are examined: (a) 20News and (b) Reddit. {Error bars indicate one standard deviation over five runs.} 
  }  
  \label{fig:len_ami}
\end{figure*}

To further assess performance with respect to document length, we analyzed topic quality across different document lengths using two datasets: 20News and Reddit.
Each dataset was divided into four groups according to word count. 
For 20News, the groups were: [1--150], [151--250], [251--400], and [401+], with approximately equal numbers of documents in each. 
For Reddit, we adopted the grouping scheme defined in the literature~\cite{Curiskis_2019}: [1--100], [101--200], [201--500], and [501+].
{Our framework and baseline methods were applied} to each document group and computed AMI and ARI scores. 
Figure~\ref{fig:len_ami} illustrates the effect of document length on AMI score (see Appendix~\ref{appendix_ari} for ARI results). 
TopiCLEAR(GMM) achieved the highest AMI and ARI scores across all document length groups. 
These findings suggest that combining SBERT embeddings with adaptive, projection-based clustering is an effective strategy for identifying high-quality topics from texts with varying linguistic styles and document lengths. 

%  追加: 計算時間の結果
{In addition, we compared the runtime of TopiCLEAR, baseline methods, and BERTopic on the AgNewsTitle dataset using 1,000--100,000 documents (Table~\ref{tab:runtime}), including a $k$-means-based implementation of TopiCLEAR.
All experiments were performed on a 2023 14-inch MacBook Pro (Apple M3 Pro, 11-core CPU, 18 GB unified memory) using the CPU only.
Even for 100,000 documents, TopiCLEAR completed within $9.4 \pm 0.8$ seconds ($k$-means) and $49.4 \pm 2.2$ seconds (GMM). At this scale, TopiCLEAR($k$-means, GMM) were faster than BERTopic, while TopiCLEAR($k$-means) achieved runtime comparable to LDA and BTM.
}

\begin{table}[t]
\centering
\caption{
{Runtime comparison. 
Results are reported as mean $\pm$ standard deviation (seconds) over five runs. The lowest mean value is shown in {\bf bold}. 
Note that all computations were performed using only the CPU and the time required for sentence embedding was excluded for BERTopic, CTM, and TopiCLEAR($k$-means and GMM). The measurement was terminated if the runtime reached two hours (7,200 seconds).}
}
\label{tab:runtime}
\vspace{0.3cm}
\begingroup
\setlength{\tabcolsep}{3.2pt}
\scriptsize
\begin{tabular}{@{}lccc@{}}
\toprule
\textbf{Method} & \textbf{1,000} & \textbf{10,000} & \textbf{100,000} \\
\midrule
LDA & 0.23 $\pm$ 0.02  & 1.47 $\pm$ 0.34 & 11.3 $\pm$ 0.6  \\[1pt]
BTM & {\bf 0.11 $\pm$ 0.00}  & {\bf 1.07 $\pm$ 0.05} & 12.1 $\pm$ 1.4 \\[1pt]
ProdLDA & 100.6 $\pm$ 6.4  & 1432 $\pm$ 154 & $>7200$\\[1pt]
ETM & 40.3 $\pm$ 3.0    &  999.7 $\pm$ 240.1 & $>7200$\\[1pt]
CTM & 141.1 $\pm$ 14.1  & 1312 $\pm$ 375 & $>7200$\\[1pt]
BERTopic & 1.29 $\pm$ 0.04 & 11.3 $\pm$ 3.5 & 74.9 $\pm$ 4.5  \\[1pt]  
TopiCLEAR($k$-means) & 2.49 $\pm$ 0.62 & 1.86 $\pm$ 0.17 & {\bf 9.40 $\pm$ 0.75}  \\[1pt]
TopiCLEAR(GMM) & 2.67 $\pm$ 0.75 & 4.06 $\pm$ 1.34 & 49.4 $\pm$ 2.2  \\
\bottomrule
\end{tabular}
\endgroup
\end{table}

\clearpage

\subsection{Ablation study}
\label{section_ablation}

We conduct an ablation study of TopiCLEAR to better understand the relative importance of the embedding model, clustering method, and preprocessing procedure. 
First, we evaluate the impact of the choice of sentence embeddings on the AMI score, which measures agreement between discovered topics and human-labeled topics (Table~\ref{tab:ablation_emb_ami}). 
We consider three embeddings: \texttt{all-MiniLM-L6-v2}, \texttt{paraphrase‑MiniLM‑L6‑v2}, and \texttt{intfloat/e5-small-v2} (Appendix~\ref{appendix_embedding} for details). 
Both \texttt{all-MiniLM-L6-v2} and \texttt{intfloat/e5-small-v2} achieve high AMI scores on the news datasets (20News and AgNewsTitle), while \texttt{all-MiniLM-L6-v2} performs best on social media datasets (Reddit and TweetTopic). 
These results indicate that embedding choice is particularly important for informal texts. Nevertheless, TopiCLEAR consistently outperforms all baselines in AMI and ARI regardless of the embedding (Appendix~\ref{appendix_ablation} for ARI results).

Second, we evaluate the impact of the clustering method on topic quality. 
Table~\ref{tab:ablation_k_ami} shows how AMI varies with the number of topics $K$ (Appendix~\ref{appendix_ablation} for ARI results). Except for TweetTopic, AMI is highest when $K$ matches its true value. 
Although the true $K$ is unknown in practice, we find that the value maximizing $C_v$ provides an approximate estimate, even though it tends to overestimate the true value.
{
Next, we evaluate the effect of Adaptive Dimensionality Reduction (ADR) by comparing the performance at the initial state with that after convergence (Table~\ref{tab:ablation_clust_ami} and Appendix~\ref{appendix_ablation} for ARI results).  
The results show that ADR consistently improves the AMI score across all datasets for both $k$-means and GMM. ADR also improves the ARI score the two formal-text datasets (20News and AGNews),  although its effect is not significant for the informal-text datasets (Reddit and TweetTopic). Overall, ADR provides a consistent benefit in terms of AMI, while its impact on ARI depends on the characteristics of the dataset. 
}  
%%  Next, we fix the number of clusters to the true value for all methods except HDBSCAN and examine the effect of clustering methods (Table~\ref{tab:ablation_clust_ami} and Appendix~\ref{appendix_ablation} for ARI results). Methods with a fixed number of clusters---$k$-means, GMM, and their ADR-enhanced versions---consistently outperform HDBSCAN in both AMI and ARI. These results indicate that specifying topic granularity ($K$) plays an important role in aligning discovered topics with human-labeled topics. In addition, incorporating ADR leads to a modest improvement for both $k$‑means and GMM, except for $k$‑means on AgNewsTitle. 

Finally, we examine the effect of preprocessing on topic quality (Table~\ref{tab:ablation_pre_ami} and Appendix~\ref{appendix_ablation} for ARI results). 
{
PCA-based preprocessing consistently improves AMI and generally improves ARI compared with using the original embeddings, and $D = 64$ achieves the highest AMI on three of the four benchmark datasets. 
Although $D = 128$ performs slightly better on some datasets, the gains are limited and dataset-dependent. Therefore, we selected $D=64$ as a practical trade-off between accuracy and efficiency.} 
%  \clearpage 

%%%%%%%%%%%%%%%
%%%  Effect of sentence embedding
%%%%%%%%%%%%%%%
\begin{table}[t]
\centering
\caption{
{Effect of sentence embedding on AMI. Results are reported as mean $\pm$ standard deviation over five runs. The highest mean value is shown in {\bf bold}. }
}
\label{tab:ablation_emb_ami}
\begingroup
\setlength{\tabcolsep}{2.5pt}
\footnotesize
\begin{tabular}{@{}lcccc@{}}
\toprule
\textbf{Embedding} & \textbf{20News} & \textbf{AgNewsTitle} & \textbf{Reddit} & \textbf{TweetTopic} \\
\midrule
\texttt{all-MiniLM-L6-v2} & \textbf{0.588 $\pm$ 0.006} & \textbf{0.489 $\pm$ 0.000} & \textbf{0.462 $\pm$ 0.000} & \textbf{0.382 $\pm$ 0.004} \\
\texttt{paraphrase-MiniLM-L6-v2} & 0.533 $\pm$ 0.007 & 0.419 $\pm$ 0.000 & 0.347 $\pm$ 0.000 & 0.306 $\pm$ 0.019 \\
\texttt{intfloat/e5-small-v2} & 0.586 $\pm$ 0.014 & 0.457 $\pm$ 0.000 & 0.323 $\pm$ 0.000 & 0.365 $\pm$ 0.001 \\
\bottomrule
\end{tabular}
\endgroup
\end{table}

%%%%%%%%%%%%%%%
%%%  Effect of the number of topics
%%%%%%%%%%%%%%%
\begin{table}[t]
\centering
\caption{
  {Effect of the number of topics $K$ on AMI and \textbf{$C_v$}. 
  Results are reported as mean $\pm$ standard deviation over five runs. The highest mean value is shown in {\bf bold}. The true number of topics is highlighted in \magenta{magenta}.} 
}
\label{tab:ablation_k_ami}
\vspace{0.3cm}
\begingroup
\setlength{\tabcolsep}{5pt}
\begin{tabular}{@{}llcc@{}}
\toprule
\textbf{Dataset} & \textbf{$K$} & \textbf{AMI} & \textbf{$C_v$} \\
\midrule
20News & 5 & 0.445 $\pm$ 0.000 & 0.514 $\pm$ 0.000 \\
 & 10 & 0.566 $\pm$ 0.005 & 0.524 $\pm$ 0.004 \\
 & \magenta{20} & \textbf{0.588 $\pm$ 0.006} & 0.552 $\pm$ 0.007 \\
 & 30 & 0.578 $\pm$ 0.004 & 0.568 $\pm$ 0.017 \\
 & 40 & 0.569 $\pm$ 0.005 & \textbf{0.584 $\pm$ 0.002} \\
\midrule
AgNewsTitle & 2 & 0.289 $\pm$ 0.000 & 0.330 $\pm$ 0.000 \\
 & \magenta{4} & \textbf{0.489 $\pm$ 0.000} & 0.351 $\pm$ 0.000 \\
 & 6 & 0.423 $\pm$ 0.000 & \textbf{0.389 $\pm$ 0.004} \\
 & 8 & 0.391 $\pm$ 0.002 & 0.355 $\pm$ 0.004 \\
 & 10 & 0.389 $\pm$ 0.008 & 0.379 $\pm$ 0.017 \\
\midrule
Reddit & 3 & 0.373 $\pm$ 0.000 & 0.437 $\pm$ 0.000 \\
 & \magenta{5} & \textbf{0.462 $\pm$ 0.000} & 0.433 $\pm$ 0.000 \\
 & 8 & 0.453 $\pm$ 0.001 & 0.434 $\pm$ 0.003 \\
 & 10 & 0.435 $\pm$ 0.000 & 0.449 $\pm$ 0.003 \\
 & 13 & 0.410 $\pm$ 0.002 & \textbf{0.452 $\pm$ 0.003} \\
\midrule
TweetTopic & 3 & \textbf{0.504 $\pm$ 0.001} & 0.434 $\pm$ 0.000 \\
 & \magenta{6} & 0.382 $\pm$ 0.004 & 0.411 $\pm$ 0.004 \\
 & 9 & 0.399 $\pm$ 0.007 & \textbf{0.438 $\pm$ 0.008} \\
 & 12 & 0.367 $\pm$ 0.013 & 0.434 $\pm$ 0.019 \\
 & 15 & 0.359 $\pm$ 0.003 & 0.422 $\pm$ 0.009 \\
\bottomrule
\end{tabular}
\endgroup
\end{table}

\begin{table}[t]
\centering
\caption{{
Effect of ADR on AMI. Results are reported as mean $\pm$ standard deviation over five runs. The highest mean value is shown in {\bf bold}. } 
}
%% The largest mean is highlighted in {\bf bold}.}}
%% {Effect of clustering methods on the AMI score. Values are reported as the mean $\pm$ sample standard deviation over five runs. The largest unrounded mean is highlighted in {\bf bold}.}}
\label{tab:ablation_clust_ami}
\begingroup
\setlength{\tabcolsep}{2.5pt}
\small
\begin{tabular}{@{}lcccc@{}}
\toprule
   & \textbf{20News} & \textbf{AgNewsTitle} & \textbf{Reddit} & \textbf{TweetTopic} \\
\midrule
    & \multicolumn{4}{c}{{\bf $k$-means} } \\
   Initial  & 0.553 $\pm$ 0.006 & 0.451 $\pm$ 0.000 & 0.407 $\pm$ 0.000 & 0.329 $\pm$ 0.001 \\
   Converged & {\bf 0.600 $\pm$ 0.008} & 0.478 $\pm$ 0.000 & 0.452 $\pm$ 0.000 & 0.377 $\pm$ 0.001 \\
   & \multicolumn{4}{c}{{\bf GMM} } \\
   Initial  & 0.562 $\pm$ 0.007 &  0.452 $\pm$ 0.000 & 0.417 $\pm$ 0.000 & 0.357 $\pm$ 0.000 \\
   Converged  & 0.588 $\pm$ 0.006 &  {\bf 0.489 $\pm$ 0.000} & {\bf 0.462 $\pm$ 0.000} & {\bf 0.382 $\pm$ 0.004} \\
%%  HDBSCAN & 0.006 $\pm$ 0.000 & 0.106 $\pm$ 0.000 & 0.162 $\pm$ 0.000 & 0.191 $\pm$ 0.000 \\
%%  $k$-means & 0.567 $\pm$ 0.017 & 0.452 $\pm$ 0.029 & 0.433 $\pm$ 0.012 & 0.348 $\pm$ 0.017 \\
%%  GMM & 0.600 $\pm$ 0.010 & \textbf{0.491 $\pm$ 0.000} & 0.454 $\pm$ 0.004 & \textbf{0.389 $\pm$ 0.011} \\
%%  ADR: $k$-means & \textbf{0.600 $\pm$ 0.008} & 0.478 $\pm$ 0.000 & 0.452 $\pm$ 0.000 & 0.377 $\pm$ 0.001 \\
%%  ADR: GMM & 0.588 $\pm$ 0.006 & 0.489 $\pm$ 0.000 & \textbf{0.462 $\pm$ 0.000} & 0.382 $\pm$ 0.004 \\
\bottomrule
\end{tabular}
\endgroup
\end{table}

%  Effect of Preprocessing
\begin{table}[t]
\centering
  \caption{{Effect of preprocessing step on AMI. Results are reported as mean $\pm$ standard deviation over five runs. 
  {\bf Bold} and \underline{underlined} values indicate the highest and second-highest mean values, respectively.}   
  }
\label{tab:ablation_pre_ami}
\begingroup
\setlength{\tabcolsep}{3.2pt}
\small
\begin{tabular}{@{}lcccc@{}}
\toprule
\textbf{Preprocessing} & \textbf{20News} & \textbf{AgNewsTitle} & \textbf{Reddit} & \textbf{TweetTopic} \\
\midrule
\textbf{PCA: $D{=}32$} & 0.577 $\pm$ 0.007 & \underline{0.488 $\pm$ 0.000} & 0.457 $\pm$ 0.000 & 0.379 $\pm$ 0.008 \\
\textbf{PCA: $D{=}64$} & \underline{0.588 $\pm$ 0.006} & \textbf{0.489 $\pm$ 0.000} & \textbf{0.462 $\pm$ 0.000} & \textbf{0.382 $\pm$ 0.004} \\
\textbf{PCA: $D{=}128$} & \textbf{0.592 $\pm$ 0.006} & 0.484 $\pm$ 0.000 & \underline{0.460 $\pm$ 0.000} & \underline{0.382 $\pm$ 0.006} \\
\textbf{None} & 0.564 $\pm$ 0.002 & 0.448 $\pm$ 0.000 & 0.411 $\pm$ 0.000 & 0.361 $\pm$ 0.000 \\
\bottomrule
\end{tabular}
\endgroup
\end{table}

\clearpage

\section{Qualitative Evaluation} 

We conduct a case study on a Twitter dataset to evaluate the proposed {framework} (TopiCLEAR) from two perspectives: (i) a comparison with human annotations and (ii) a qualitative analysis without ground-truth labels. 
Here, we compare the topics identified from the TweetTopic dataset by two methods: TopiCLEAR and LDA, a widely used topic discovery method. 
Table~\ref{top_wprds} presents the top 20 words for each topic identified by both methods using the same number of topics ($K{=}\,6$). 
To align topics across the two methods, we apply a greedy matching algorithm based on cosine similarity. The low similarity scores ($\leq 0.14$) suggest that the topics obtained by TopiCLEAR differ from those obtained by LDA. 
We further focus on the ``pop culture'' topic, as identified by human annotators. The representative 20 words associated with this topic are highlighted in bold. These words were selected by computing the delta TF-IDF~\cite{Martineau_2009} between the ``pop culture'' topic and the average across all topics. 
In TopiCLEAR, most of the bolded words appear prominently in Topic 5, while the remaining top words in Topic 5 (e.g., ``URLs'' and ``usernames'') are general rather than topic-specific. 
By contrast, in LDA, the bolded words are scattered across three topics (Topics 4, 5, and 6), as shown in the right panel of Table~\ref{top_wprds}. 
These observations suggest that TopiCLEAR more effectively captures the ``pop culture'' topic identified by annotators, whereas LDA does not isolate this topic clearly.

\begin{figure}[t]
  \centering
  \includegraphics[width=15cm]{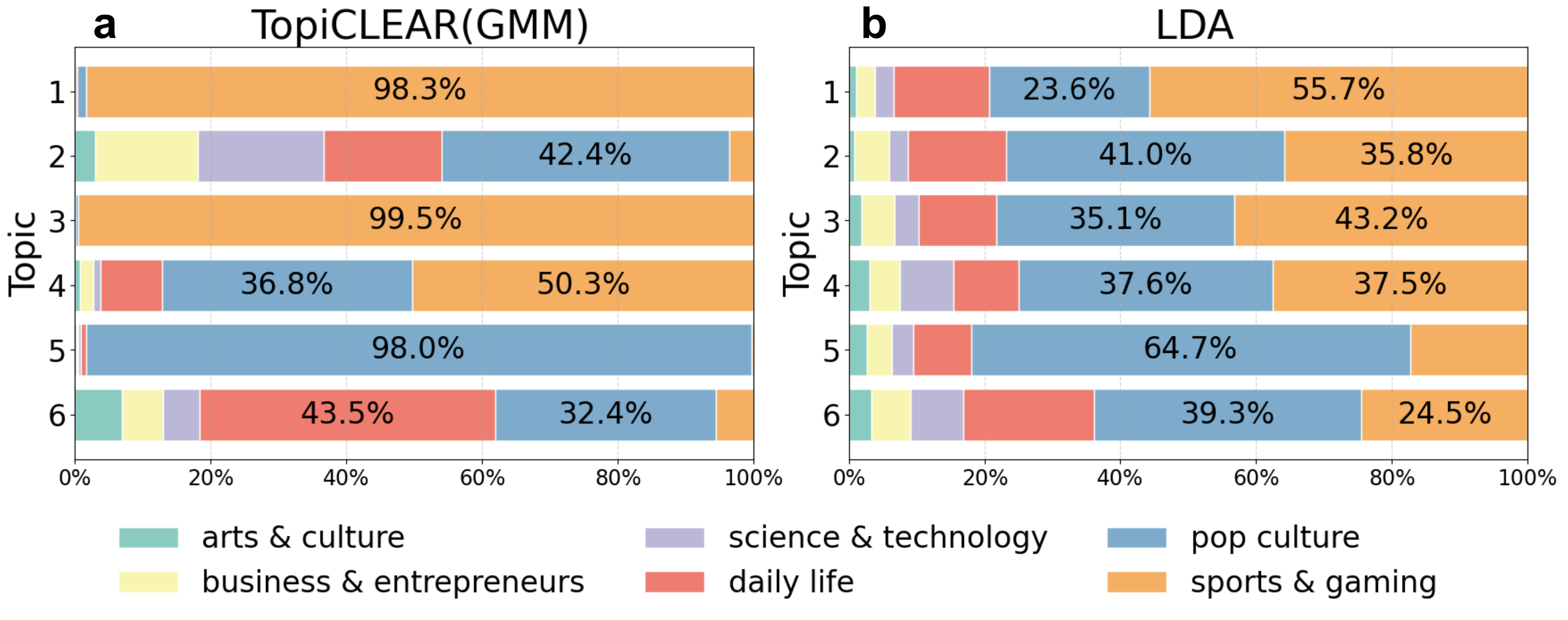}
  \caption{
  Composition of human-annotated topics in topics discovered by (a) TopiCLEAR(GMM) and (b) LDA on TweetTopic. Percentages are shown only for values $\geq 20 \%$.}   
  \label{fig:qualitative}
\end{figure}

To assess topic interpretability, we examine representative documents for each topic identified by TopiCLEAR and LDA, and compare the topic assignments with human-annotated labels. 
Topics 1 and 3 contain a large number of tweets related to sports, while Topic 5 contains numerous music-related tweets.  
According to the annotators' classification, 98.3\% and 99.5\% of the tweets in Topics 1 and 3, respectively, were assigned to ``sports \& gaming,'' while 98.0\% of the tweets in Topic 5 were assigned to ``pop culture'' (Figure~\ref{fig:qualitative}).
Additionally, Topic 1 focused primarily on internationally popular sports such as football (Premier League) and UFC, while Topic 3 centered on sports popular in the United States, such as the NBA and NFL. 
Topic 6 consisted mainly of personal messages celebrating holidays such as Easter or expressing greetings to family and other loved ones. Although the annotators classified Topic 6 as a mixture of multiple topics, its content remained interpretable. 
\clearpage

\begin{table}[t]
\centering
\caption{
Top 20 words for each topic identified by TopiCLEAR(GMM) (left) and LDA (right) on the TweetTopic dataset. 
Words are ordered by frequency for TopiCLEAR and by $P(w|z)$ for LDA, where $w$ and $z$ denote a word and a topic, respectively. 
The 20 most specific words for the ``pop culture'' topic, identified using delta TF-IDF~\cite{Martineau_2009}, are shown in bold. Sim denotes the cosine similarity between topics.
}
\vspace{0.3cm}
\label{top_wprds}
\begin{tabular}{|c|p{5.4cm}|p{5.4cm}|c|}
\hline
Topic & TopiCLEAR(GMM) & LDA & Sim \\ \hline
1 & username, url, via, ufc, league, win, game, fight, vs, city, fc, manchester, final, time, match, back, today, football, first, united
  & username, url, league, game, football, team, thank, happy, please, day, city, win, time, love, stay, \textbf{bts\_twt}, today, state, still, make 
  & 0.140 \\ \hline
2 & url, username, via, news, new, change, please, time, climate, trump, us, \textbf{movie}, like, people, get, one, make, bad, star, watch
  & username, love, like, one, get, url, today, time, every, really, year, got, please, back, game, ever, top, vs, good, need
  & 0.119 \\ \hline
3 & username, url, game, nfl, win, state, team, football, via, chiefs, first, season, one, vs, today, ohio, bowl, series, get, browns
  & username, url, game, see, morning, great, make, new, sure, \textbf{video}, listen, well, going, win, let, hope, pretty, made, team, get 
  & 0.079 \\ \hline
4 & username, url, via, new, game, time, get, day, one, see, love, live, today, \textbf{youtube}, great, night, year, team, back, win
  & url, username, via, new, \textbf{bts\_twt}, ufc, \textbf{permission}, \textbf{dance}, \textbf{spotify}, watch, \textbf{bts}, get, vs, \textbf{youtube}, first, year, butter, trump, like, live 
  & 0.067 \\ \hline
5 & url, username, \textbf{music}, new, \textbf{video}, via, \textbf{youtube}, \textbf{album}, \textbf{song}, \textbf{official}, love, \textbf{harry}, live, \textbf{bts\_twt}, \textbf{styles}, \textbf{spotify}, listen, \textbf{nowplaying}, \textbf{bts}, \textbf{watermelon}
  & url, via, \textbf{music}, \textbf{youtube}, \textbf{video}, new, \textbf{official}, username, day, live, check, apple, \textbf{album}, news, love, world, happy, \textbf{song}, father, lyrics 
  & 0.036 \\ \hline
6 & username, url, love, day, happy, via, new, year, thank, time, please, see, one, like, great, today, us, much, good, family
  & username, url, new, via, happy, \textbf{music}, day, time, love, \textbf{video}, \textbf{youtube}, first, one, news, \textbf{song}, great, \textbf{album}, get, thanks, us 
  & 0.031 \\ \hline
\end{tabular}
\vspace{1cm}
\end{table}
\clearpage

Using the same procedure, we examine the topics identified by LDA.
Topic 1 contains a mixture of sports-related and music-related tweets, with top words including the music-related term ``bts\_twt.''
Similarly, Topics 3 and 5 also consist of mixed sports and music content.
According to annotator labels, all six LDA topics contain a mixture of tweets spanning multiple categories, including ``pop culture,'' ``daily life,'' and ``sports \& gaming'' (Figure~\ref{fig:qualitative}).

%%%%%%%%%%%%%%%%%%%%%%%%%%%%%%
%%%%%  Sub-topic level analysis
%%%%%%%%%%%%%%%%%%%%%%%%%%%%%%
Next, we examine whether TopiCLEAR can discover fine-grained sub-topic structures within a coarse topic category.
Here, we focus on the ``sports \& gaming'' category, which is one of the largest categories (2,483 tweets) and contains diverse subtopics such as different sports competitions and video games.
We compare the topics identified from tweets in this category using two methods: TopiCLEAR and LDA. For a fair comparison, we fix the number of topics to $K=4$ for both methods.
Table~\ref{top_words_2} presents the top 20 words for each topic identified by the two methods. Appendix~\ref{appendix_output} provides representative documents for each topic. 
As in Table~\ref{top_wprds}, topics are aligned based on cosine similarity.
The low similarity scores ($\leq 0.12$) suggest that the topics discovered by TopiCLEAR differ substantially from those obtained by LDA. 
\newpage

%  Interpretation of TopiCLEAR 
TopiCLEAR discovers semantically coherent subtopics corresponding to specific sports competitions (Table~\ref{top_words_2} and Appendix~\ref{appendix_output}).
Topic A (1,052 tweets) contains terms such as ``NFL'', ``Chiefs'', ``Browns'', and ``Super Bowl'', indicating American football.
Topic B (514 tweets) includes ``FC'', ``Manchester'', ``Chelsea'', and ``Premier'', corresponding to Association football (soccor or football).
Topic C (261 tweets) is characterized by ``UFC'', ``fight'', ``MMA'', and ``boxing'', representing martial arts.
In contrast, Topic D (656 tweets) contains generic terms such as ``game'', ``win'', ``live'', and ``today'' and does not correspond to a specific sports competition.
%  Interpretation of LDA
By contrast, LDA does not clearly separate these subtopics; for example, the term ``UFC'' appears among the top words of multiple topics, indicating that martial-arts-related tweets are split across topics. 
%  Appendix E
Additionally, the content of each topic identified by TopiCLEAR is straightforward to interpret from representative documents, whereas this is less clear for LDA (Appendix~\ref{appendix_output}). 
To quantitatively support these observations, we examine topic assignments for tweets containing representative terms.
Among 116 tweets containing ``NFL'', TopiCLEAR assigns 111 tweets (96\%) to Topic A, whereas LDA assigns 44 tweets (38\%) as Topic A, which had the highest proportion. 
Similarly, among 147 tweets containing ``FC'', ``Manchester'', ``Chelsea'', or ``Barcelona'', TopiCLEAR assigns 143 tweets (97\%) to Topic B, whereas LDA assigns 75 tweets (51\%) to Topic B, which had the highest proportion. 
Finally, among 123 tweets containing ``UFC'', TopiCLEAR assigns 120 tweets (98\%) to Topic C, whereas LDA splits these tweets between Topic A (46\%) and Topic C (45\%).

Overall, these results demonstrate that TopiCLEAR more effectively isolates both coarse topics and fine-grained subtopic structures than LDA. At the coarse level, TopiCLEAR discovers topics that closely match annotator-defined categories or can be readily interpreted through representative documents. In contrast, LDA topics consistently mix multiple categories. At the fine-grained level, TopiCLEAR recovers subtopics corresponding to American football, Association football, and martial arts, while LDA disperses salient terms across topics, making interpretation more challenging.

\begin{table}[t]
\centering
\caption{
Top 20 words for each topic identified by TopiCLEAR(GMM) (left) and LDA (right) in the ``sports \& gaming'' subset of TweetTopic. 
As in Table~\ref{top_wprds}, words are ordered by frequency for TopiCLEAR and by $P(w|z)$ for LDA. 
Sim denotes the cosine similarity between topics.
}
\vspace{0.3cm}
\label{top_words_2}
\begin{tabular}{|c|p{5.4cm}|p{5.4cm}|c|}
\hline
Topic & TopiCLEAR(GMM) & LDA & Sim \\ \hline
A & username, url, game, nfl, win, state, via, football, team, chiefs, today, bowl, one, get, first, ohio, browns, super, new, series
  & url, username, via, game, league, one, first, final, get, cup, great, go, sports, win, ufc, best, see, new, next, fc
  & 0.118 \\ \hline
B & username, url, league, fc, city, manchester, game, final, cup, club, football, season, united, team, fans, chelsea, via, win, premier, match
  & username, url, football, city, team, chelsea, league, state, fc, win, manchester, season, via, united, today, final, fans, teams, game, ohio
  & 0.085 \\ \hline
C & url, ufc, via, fight, username, wilder, fury, tyson, deontay, vs, mma, 245, anthony, joshua, boxing, junkie, win, live, ruiz, 251
  & username, url, game, via, vs, team, football, ufc, season, new, time, last, back, one, play, sports, us, see, fc, win
  & 0.056 \\ \hline
D & username, url, game, via, win, first, today, new, good, time, one, get, back, vs, games, team, great, day, live, year
  & username, url, game, team, today, night, good, back, match, first, win, next, get, second, games, like, year, day, take, come
  & 0.035 \\ \hline
\end{tabular}
\vspace{1cm}
\end{table}
\clearpage

\section{Conclusion}   

In this study, we {present} TopiCLEAR, a simple framework for topic discovery that {constructs} a low-dimensional topic space from document embeddings through iterative clustering with adaptive dimensionality reduction. 
This {framework} is motivated by the hypothesis that human-interpretable topics correspond to low-dimensional geometric structures in embedding spaces.
We evaluated the proposed {framework} on four benchmark datasets covering both formal and informal texts. 
TopiCLEAR(GMM) achieves consistently higher AMI and ARI scores than the baselines, indicating that it identifies topics that align with human classifications, particularly for short and informal texts (Tables~\ref{tab:score_AMI} and \ref{tab:score_ARI}). 
{
Our quantitative evaluation relies on benchmark datasets with human-annotated topic labels. 
While such datasets enable reproducible comparisons, they may favor methods that align with the annotation scheme and may not fully capture alternative topic granularities or interpretations. 
We therefore complemented the quantitative evaluation with a qualitative analysis on the TweetTopic dataset (Sec.~6), comparing the topic and sub-topic structures discovered by TopiCLEAR and LDA. 
This analysis provides complementary evidence of topic interpretability beyond AMI and ARI.
} 
%%  demonstrated that TopiCLEAR produces more interpretable topics than Latent Dirichlet Allocation (LDA), recovering human-annotated topics and identifying coherent sub-topic structures.
% 
Overall, these findings highlight the effectiveness of low-dimensional topic representations for discovering interpretable topics from diverse text corpora.

Despite these promising results, TopiCLEAR has several limitations. First, the user must specify the topic granularity, i.e., the number of topics $K$. 
{The ablation study suggests that  the topic coherence metric $C_v$ tends to overestimate the number of topics (Table~\ref{tab:ablation_k_ami}). Therefore, we do not recommend relying solely on the coherence metric. Instead, we suggest using $C_v$ as an initial guide for selecting $K$ and then refining the choice through qualitative inspection (e.g., reading) of the resulting topics.} 
Another limitation is the assumption of a linear subspace as the topic space, which may hinder topic discovery performance. 
Future work includes developing methods to automatically determine the number of topics, extending the framework to capture hierarchical topic structures, and exploring clustering methods on curved spaces.  

\section*{Declaration of generative AI and AI-assisted technologies in the manuscript preparation process} 
During the manuscript preparation, the authors used Microsoft Copilot for proofreading the manuscript. The authors reviewed and edited the output as needed and take full responsibility for the content of the published article.

\section*{Acknowledgements}
We thank Yuka Takedomi for the stimulating discussions. 
This study was supported by the World-leading Innovative Graduate Study Program in Proactive Environmental Studies (WINGS-PES), the University of Tokyo, to T.Y., and 
JSPS KAKENHI (Nos. JP21H03559, JP22H03695, and JP23K24950), JST FOREST (No. JPMJFR232O), and AMED (No. JP223fa627001) to R.K.

\bibliographystyle{unsrt}  
\bibliography{references2025.bib}

@article{blei_latent_2003,
author = {Blei, David M. and Ng, Andrew Y. and Jordan, Michael I.},
title = {Latent Dirichlet allocation},
year = {2003},
volume = {3},
issn = {1532-4435},
journal = {Journal of Machine Learning Research},
pages = {993--1022},
url     = {https://www.jmlr.org/papers/v3/blei03a.html},
}

@article{blei2012probabilistic,
  author  = {David M. Blei},
  title   = {Probabilistic topic models},
  journal = {{Communications of the ACM}},
  volume  = {55},
  number  = {4},
  pages   = {77--84},
  year    = {2012},
  doi     = {10.1145/2133806.2133826},
}

@article{kobayashi_evolution_2022,
  title = {Evolution of public opinion on COVID-19 vaccination in Japan: Large-scale {Twitter} data analysis},
  author = {Kobayashi, Ryota and Takedomi, Yuka and Nakayama, Yuri and Suda, Towa and Uno, Takeaki and Hashimoto, Takako and Toyoda, Masashi and Yoshinaga, Naoki and Kitsuregawa, Masaru and Rocha, Luis E. C.},
  year = {2022},
  journal = {Journal of Medical Internet Research},
  volume = {24},
  number = {12},
  pages = {e41928},
  publisher = {JMIR Publications Inc., Toronto, Canada},
  doi = {10.2196/41928},
  url = {https://www.jmir.org/2022/12/e41928},
  urldate = {2023-01-08},
}

@article{pedregosa_scikit-learn_2011,
  author  = {Fabian Pedregosa and Ga{{\"e}}l Varoquaux and Alexandre Gramfort and Vincent Michel and Bertrand Thirion and Olivier Grisel and Mathieu Blondel and Peter Prettenhofer and Ron Weiss and Vincent Dubourg and others},
  title   = {{Scikit-learn: Machine learning in Python}},
  journal = {Journal of Machine Learning Research},
  year    = {2011},
  volume  = {12},
  pages   = {2825--2830},
  url     = {https://jmlr.org/papers/v12/pedregosa11a.html}
}

@inproceedings{reimers_sentence-bert_2019,
    title = "Sentence-{BERT}: Sentence embeddings using {S}iamese {BERT}-networks",
    author = "Reimers, Nils  and
      Gurevych, Iryna",
    booktitle = "Proceedings of the 2019 Conference on Empirical Methods in Natural Language Processing and the 9th International Joint Conference on Natural Language Processing",
    year = "2019",
    location = "Hong Kong, China",
    pages = "3982--3992",   
    doi = "10.18653/v1/D19-1410",    
}

@inproceedings{vinh_information_2009,
  title = {Information theoretic measures for clusterings comparison: Is a correction for chance necessary?\nodot},
  shorttitle = {Information Theoretic Measures for Clusterings Comparison},
  booktitle = {Proceedings of the 26th International Conference on Machine Learning (ICML '09)},
  author = {Vinh, Nguyen Xuan and Epps, Julien and Bailey, James},
  year = {2009},
  pages = {1073--1080},  
  doi = {10.1145/1553374.1553511},
}

@inproceedings{zhang_character-level_2015,
  author    = {Zhang, Xiang and Zhao, Junbo and LeCun, Yann},
  title     = {Character-level convolutional networks for text classification},
  booktitle = {Advances in Neural Information Processing Systems 28 (NIPS'15)},
  pages     = {649--657},
  year      = {2015},
}

@inproceedings{Yan_2013,
  author    = {Xiaohui Yan and Jiafeng Guo and Yanyan Lan and Xueqi Cheng},
  title     = {A biterm topic model for short texts},
  booktitle = {Proceedings of the 22nd International Conference on World Wide Web (WWW 13)},
  year      = {2013},
  pages     = {1445--1456},  
  doi       = {10.1145/2488388.2488514},
}

@inproceedings{Srivastava_2017,
  author    = {Akash Srivastava and Charles A. Sutton},
  title     = {Autoencoding variational inference for topic models},
  booktitle = {Proceedings of the 5th International Conference on Learning Representations ({ICLR} 2017)},
  year      = {2017},
  url       = {https://openreview.net/forum?id=BybtVK9lg}
}

@article{Dieng_2020,
    title = "Topic modeling in embedding spaces",
    author = "Dieng, Adji B.  and
      Ruiz, Francisco J. R.  and
      Blei, David M.",
    journal = "Transactions of the Association for Computational Linguistics",
    volume = "8",
    year = "2020",
    address = "Cambridge, MA",
    publisher = "MIT Press",
    url = "https://aclanthology.org/2020.tacl-1.29/",
    doi = "10.1162/tacl_a_00325",
    pages = "439--453",
}

@inproceedings{bianchi_2021,
    title = "Cross-lingual contextualized topic models with zero-shot learning",
    author = "Bianchi, Federico  and
      Terragni, Silvia  and
      Hovy, Dirk  and
      Nozza, Debora  and
      Fersini, Elisabetta",
    booktitle = "Proceedings of the 16th Conference of the European Chapter of the Association for Computational Linguistics: Main Volume (EACL 2021)",
    year = "2021",
    pages = "1676--1683",
    location = "Online",    
    doi = "10.18653/v1/2021.eacl-main.143", 
}

@inproceedings{antypas_2022,
    title = "{Twitter} topic classification",
    author = "Antypas, Dimosthenis  and
      Ushio, Asahi  and
      Camacho-Collados, Jose  and
      Silva, Vítor  and
      Neves, Leonardo  and
      Barbieri, Francesco",
    booktitle = "Proceedings of the 29th International Conference on Computational Linguistics (COLING 2022)",
    year = "2022",     
    pages = "3386--3400",
    doi       = {10.18653/v1/2022.coling-1.299},
}

@article{Curiskis_2019,
author={Curiskis, Stephan A. and Drake, Barry and Osborn, Thomas R. and Kennedy, Paul J.},
year = {2020},
title = {An evaluation of document clustering and topic modelling in two online social networks: Twitter and Reddit},
volume = {57},
number = {2},
issn = {0306-4573},
journal = {Information Processing \& Management},
pages   = {102034},
doi = {10.1016/j.ipm.2019.04.002}
}

@inproceedings{Ding_2007,
  author = {Ding, Chris and Li, Tao},
  title = {Adaptive dimension reduction using discriminant analysis and K-means clustering},
  year = {2007},   
  doi = {10.1145/1273496.1273562},
  booktitle = {Proceedings of the 24th International Conference on Machine Learning (ICML '07)},
  pages = {521-528},
}

@book{fukunaga_1990,
  author = {Keinosuke Fukunaga},
  title = {Introduction to statistical pattern recognition},
  edition = {2nd},
  year = {1990},
  publisher = {Academic Press},
  address = {San Diego, CA, USA},
  isbn = {9780122698514}
}

@inproceedings{roder_2015,
  author = {R\"{o}der, Michael and Both, Andreas and Hinneburg, Alexander},
  title = {Exploring the space of topic coherence measures},
  booktitle={Proceedings of the eighth ACM international conference on Web {S}earch and {D}ata {M}ining (WSDM 2015)},
  year = {2015},
  pages = {399-408},  
  doi = {10.1145/2684822.2685324},  
}

@inproceedings{Newman_2010,
  author = {Newman, David and Lau, Jey Han and Grieser, Karl and Baldwin, Timothy},
  title = {Automatic evaluation of topic coherence}, 
  booktitle = {Proceedings of the 2010 Human Language Technologies conference of the North American chapter of the Association for Computational Linguistics (NAACL HLT 2010)},
  year = {2010},
  location = {Los Angeles, CA, USA},
  pages = {100--108},  
  doi       = {10.3115/1857999.1858011}
}

@inproceedings{Hoyle_2021,
  author = {Hoyle, Alexander and Goel, Pranav and Peskov, Denis and Hian-Cheong, Andrew and {Boyd-Graber}, Jordan and Resnik, Philip},
  title = {Is automated topic model evaluation broken? The incoherence of coherence},
  booktitle = {Advances in Neural Information Processing Systems 34 (NeurIPS 2021)},
  year = {2021},
  pages     = {2018--2033},  
  url       = {https://proceedings.neurips.cc/paper/2021/hash/0f83556a305d789b1d71815e8ea4f4b0-Abstract.html}
}

@inproceedings{Wang_2020,
  author = {Wang, Wenhui and Wei, Furu and Dong, Li and Bao, Hangbo and Yang, Nan and Zhou, Ming},
  title     = {{MiniLM}: Deep self-attention distillation for task-agnostic compression of pre-trained transformers},
  year = {2020},
  booktitle = {Advances in Neural Information Processing Systems 33 (NIPS'20)},  
  pages     = {5776--5788},
}

@inproceedings{Aletras_2013,
    title = "Evaluating topic coherence using distributional semantics",
    author = "Aletras, Nikolaos  and
      Stevenson, Mark",
    booktitle = {Proceedings of the 10th International Conference on Computational Semantics},
    year = "2013",
    location = "Potsdam, Germany",    
    pages = "13--22",
    url = "https://aclanthology.org/W13-0102"
}

@inproceedings{doogan_2021,
  author    = {Doogan, Caitlin and Buntine, Wray},
  title     = {Topic model or topic twaddle? Re-evaluating semantic interpretability measures},
  booktitle = {Proceedings of the 2021 Conference of the North American Chapter of the Association for Computational Linguistics: Human Language Technologies (NAACL 2021)},
  year      = {2021},
  location = {Online},  
  pages     = {3824--3848},
  doi       = {10.18653/v1/2021.naacl-main.300},  
}

@ARTICLE{Qiang_2022,
author={Qiang, Jipeng and Qian, Zhenyu and Li, Yun and Yuan, Yunhao and Wu, Xindong},
journal={IEEE Transactions on Knowledge and Data Engineering},
title={Short text topic modeling techniques, applications, and performance: A survey},
year={2022},
volume={34},
number={3},
ISSN={1558-2191},
pages={1427--1445},
doi={10.1109/TKDE.2020.2992485},
publisher={IEEE Computer Society},
address={Los Alamitos, CA, USA}
}

@article{Hubert_1985,
  author    = {Lawrence Hubert and Phipps Arabie},
  title     = {Comparing partitions},
  journal   = {Journal of Classification},
  year      = {1985},
  volume    = {2},
  number    = {1},
  pages     = {193--218},
  doi       = {10.1007/BF01908075},
  url       = {https://doi.org/10.1007/BF01908075},
  issn      = {1432-1343},
}

@article{Lim_2024,
    author = {Lim, Jia Peng and Lauw, Hady W.},
    title = {Aligning human and computational coherence evaluations},
    journal = {Computational Linguistics},
    volume = {50},
    number = {3},
    pages = {893--952},
    year = {2024},
    issn = {0891-2017},
    doi = {10.1162/coli\_a\_00518},
}

@inproceedings{Angelov_2020,
    title = "Topic modeling: Contextual token embeddings are all you need",
    author = "Angelov, Dimo  and
      Inkpen, Diana",
  booktitle = {Findings of the Association for Computational Linguistics: {EMNLP} 2024},
  year      = {2024},
  location  = {Miami, FL, USA},  
  pages     = {13528--13539},
  doi       = {10.18653/v1/2024.findings-emnlp.790},
}

@article{hemmatian2019survey,
  title={A survey on classification techniques for opinion mining and sentiment analysis},
  author={Hemmatian, Fatemeh and Sohrabi, Mohammad Karim},
  journal={Artificial intelligence review},
  volume={52},
  number={3},
  pages={1495--1545},
  year={2019},
  publisher={Springer}
}

@article{walter2021strategy,
  title={Strategy framing in news coverage and electoral success: an analysis of topic model networks approach},
  author={Walter, Dror and Ophir, Yotam},
  journal={Political Communication},
  volume={38},
  number={6},
  pages={707--730},
  year={2021},
  publisher={Taylor \& Francis}
}

@article{wang2022text,
  title={Text embeddings by weakly-supervised contrastive pre-training},
  author={Wang, Liang and Yang, Nan and Huang, Xiaolong and Jiao, Binxing and Yang, Linjun and Jiang, Daxin and Majumder, Rangan and Wei, Furu},
  journal={Preprint, arXiv:2212.03533},
  year={2022}
}

@article{yau2014clustering,
  title={Clustering scientific documents with topic modeling},
  author={Yau, Chyi-Kwei and Porter, Alan and Newman, Nils and Suominen, Arho},
  journal={Scientometrics},
  volume={100},
  number={3},
  pages={767--786},
  year={2014},
  publisher={Springer}
}

@article{Campello_2015,
  author   = {Campello, Ricardo J. G. B. and Moulavi, Davoud and Zimek, Arthur and Sander, J{\"o}rg},
  title    = {Hierarchical density estimates for data clustering, visualization, and outlier detection},
  journal  = {ACM Transactions on Knowledge Discovery from Data},
  year     = {2015},
  volume   = {10},
  number   = {1},
  articleno= {5},
  pages = {1--51},
  doi      = {10.1145/2733381}
}

@inproceedings{Terragni_2021,
  author    = {Terragni, Silvia and Fersini, Elisabetta and Galuzzi, Bruno Giovanni and Tropeano, Pietro and Candelieri, Antonio},
  title     = {{OCTIS}: Comparing and optimizing topic models is simple!\nodot},
  year      = {2021},
  isbn      = {978-1-954085-05-3},  
  booktitle = {Proceedings of the 16th Conference of the European Chapter of the Association for Computational Linguistics: System Demonstrations},
  pages     = {263--270},
  doi       = {10.18653/v1/2021.eacl-demos.31}
}

@inproceedings{rehurek_2010,
      title = {{Software framework for topic modelling with large corpora}},
      author = {Radim {\v R}eh{\r u}{\v r}ek and Petr Sojka},
      booktitle = {{Proceedings of the LREC 2010 Workshop on New
           Challenges for NLP Frameworks}},
      pages = {45--50},
      year = {2010},
      publisher = {ELRA},
}

@inproceedings{Lang_1995,
  title = {NewsWeeder: Learning to filter netnews},
  booktitle = {Proceedings of the 12th International Conference on Machine Learning (ICML '95)},   
  pages = {331--339},
  year = {1995},
  doi = {10.1016/B978-1-55860-377-6.50048-7},
  author = {Ken Lang},
}

@ARTICLE{Frenay_2014,
  author={Fr{é}nay, Beno{î}t and Verleysen, Michel},
  journal={IEEE Transactions on Neural Networks and Learning Systems}, 
  title={Classification in the presence of label noise: A survey}, 
  year={2014},
  volume={25},
  number={5},
  pages={845--869},
  doi={10.1109/TNNLS.2013.2292894}
}

@inproceedings{Martineau_2009,
  author    = {Justin Martineau and Tim Finin},
  title     = {Delta {TFIDF}: An improved feature space for sentiment analysis},
  booktitle = {Proceedings of the International AAAI Conference on Web and Social Media (ICWSM' 09)},
  year      = {2009},
  location  = {San Jose, CA, USA},
  pages     = {258--261}, 
  doi       = {10.1609/icwsm.v3i1.13979},
}

@article{Maarten_2022,  
  title     = {BERTopic: Neural topic modeling with a class-based TF-IDF procedure},
  author    = {Maarten Grootendorst},
  year      = {2022},
  journal={Preprint, arXiv: 2203.05794}
}

@article{miyazaki-2024,
  author  = {Miyazaki, Kunihiro and others},
  title   = {Public perception of generative {AI} on {Twitter}: an empirical study based on occupation and usage},
  journal = {EPJ Data Science},
  year    = {2024},
  volume  = {13},
  articleno = {2},
  doi     = {10.1140/epjds/s13688-023-00445-y}
}

@article{yaqub-2017,
  author  = {Yaqub, Ussama and Chun, Soon Ae and Atluri, Vijayalakshmi and Vaidya, Jaideep},
  title   = {Analysis of political discourse on {Twitter} in the context of the 2016 {US} presidential elections},
  journal = {Government Information Quarterly},
  year    = {2017},
  volume  = {34},
  number  = {4},
  pages   = {613--626},
  doi     = {10.1016/j.giq.2017.11.001}
}

@article{vayansky_2020,
  author  = {Vayansky, Ike and Kumar, Sathish A. P.},
  title   = {A review of topic modeling methods},
  journal = {Information Systems},
  year    = {2020},
  volume  = {94},
  pages   = {101582},
  doi     = {10.1016/j.is.2020.101582}
}

@inproceedings{zhao2021topic,
  author={Zhao, He and Phung, Dinh and Huynh, Viet and Jin, Yuan and Du, Lan and Buntine, Wray},
  title={Topic modelling meets deep neural networks: A survey},  
  booktitle = {Proceedings of the Thirtieth International Joint Conference on Artificial Intelligence (IJCAI-21)},   
  year      = {2021},
  pages     = {4713--4720},
  doi       = {10.24963/ijcai.2021/638}
}

@inproceedings{qiang2017topic,
  title={Topic modeling over short texts by incorporating word embeddings},
  author={Qiang, Jipeng and Chen, Ping and Wang, Tong and Wu, Xindong},
  booktitle={Pacific-Asia Conference on Knowledge Discovery and Data Mining (PAKDD 2017)},
  pages={363--374},
  year={2017}  
}

@inproceedings{liu2025llm,
  title={LLM-guided semantic-aware clustering for topic modeling},
  author={Liu, Jianghan and Shang, Ziyu and Ke, Wenjun and Wang, Peng and Luo, Zhizhao and Liu, Jiajun and Li, Guozheng and Li, Yining},
  booktitle={Proceedings of the 63rd Annual Meeting of the Association for Computational Linguistics (Volume 1: Long Papers)},
  pages={18420--18435},
  year={2025}
}

@inproceedings{pham2024topicgpt,
  title={TopicGPT: A prompt-based topic modeling framework},
  author={Pham, Chau Minh and Hoyle, Alexander and Sun, Simeng and Resnik, Philip and Iyyer, Mohit},
  booktitle={Proceedings of the 2024 Conference of the North American Chapter of the Association for Computational Linguistics: Human Language Technologies (Volume 1: Long Papers)},
  pages={2956--2984},
  year={2024}
}

@article{yang2025enhancing,
  title={Enhancing topic coherence and diversity in document embeddings using LLMs: A focus on BERTopic},
  author={Yang, Chibok and Kim, Yangsok},
  journal={Expert Systems with Applications},
  volume={281},
  pages={127517},
  year={2025},
  publisher={Elsevier}
}

@inproceedings{razavi2025benchmarking,
  title={Benchmarking prompt sensitivity in large language models},
  author={Razavi, Amirhossein and Soltangheis, Mina and Arabzadeh, Negar and Salamat, Sara and Zihayat, Morteza and Bagheri, Ebrahim},
  booktitle={European Conference on Information Retrieval},
  pages={303--313},
  year={2025},
  organization={Springer}
}
\clearpage

\begin{appendices}

\section{Effect of label noise on evaluation metrics}
\label{appendix_coherence}

We analyze how label noise affects existing evaluation metrics to assess their reliability. Label noise randomly perturbs a subset of topic assignments, making clustering results increasingly resemble random partitions as the noise level grows. 
Specifically, we adopted a uniform label noise model~\cite{Frenay_2014}. 
The topic assignment for document $i$ is replaced by a uniform distribution over $\{1, \cdots, K\}$ (where $K$ is the number of topics) with probability $p_n$, and it is left unchanged with probability $1-p_n$. 
Under this model, the expected proportion of labels that match before and after adding noise is $1- \frac{K-1}{K} p_n$. As the noise level $p_n$ increases, the expected agreement between original and noisy labels decreases, leading to degraded clustering performance.

\begin{figure}[b]
  \centering
  \includegraphics[width= 14cm]{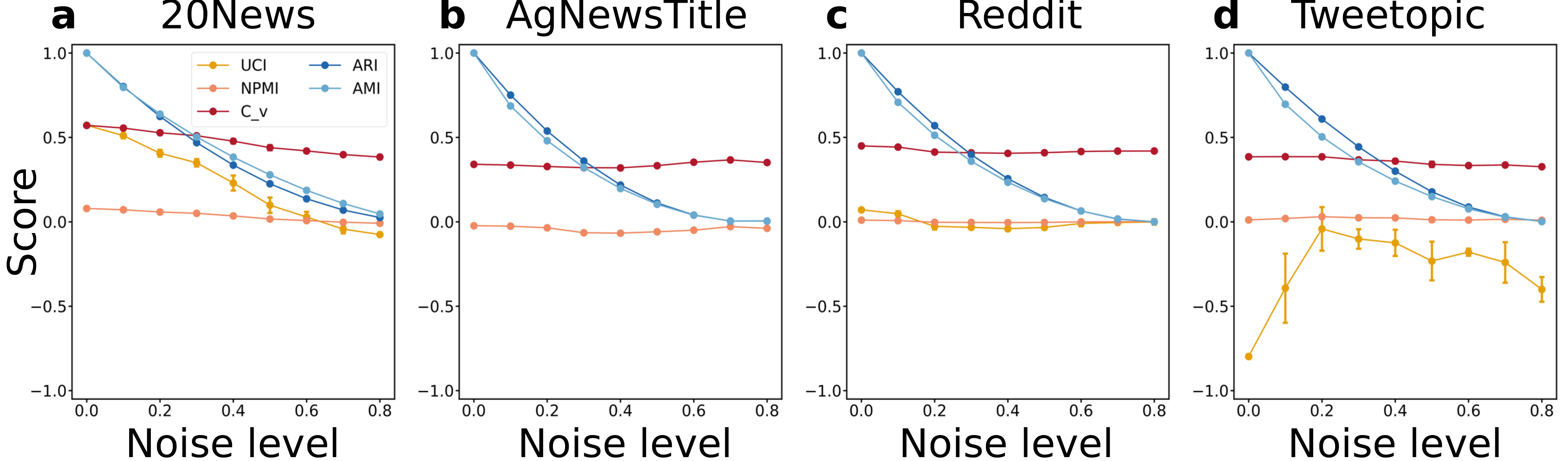}   
  \vspace{0.2cm}
  \caption{Effect of label noise on topic-discovery evaluation metrics. 
  Five metrics ($C_{\rm UCI}$, $C_{\rm NPMI}$, $C_v$; AMI and ARI) were computed after introducing label noise into the ground-truth labels. The noise level $p_n$ was increased from 0 to 0.8. 
  {Error bars indicate one standard deviation over five runs.} 
  Note that the UCI coherence values were not plotted in panel (b), because they were too small to be visible.}
  \label{figure_cv}
\end{figure}

We examined five metrics, i.e., two clustering metrics (AMI and ARI) and three coherence metrics widely used to evaluate topic models. 
The coherence metrics quantify the extent to which the most frequent words within a topic (the top $n$ words) form a coherent concept. A higher metric indicates that the topics are more interpretable. 
We adopt three topic coherence metrics: UCI~\cite{Newman_2010}, Normalized PMI (NPMI)~\cite{Aletras_2013}, and $C_v$~\cite{roder_2015}. 
UCI~$C_{\rm UCI}$ and NPMI~$C_{\rm NPMI}$ were computed using the \texttt{CoherenceModel} from Gensim library~\cite{rehurek_2010}, with a sliding window size of 10 and the top $n = 10$ words per topic. 
$C_v$ was also computed using the \texttt{CoherenceModel} from Gensim, with a sliding window size of 110 and the top $n = 10$ words per topic.

Figure~\ref{figure_cv} shows the effect of the noise level $p_n$ on evaluation metrics. 
For all datasets, AMI and ARI decreased as the noise level increased. In contrast, NPMI and $C_v$ remained nearly constant regardless of the noise level. 
Furthermore, we calculated the Spearman rank correlation coefficient $\rho$ between the noise level $p_n$ and the evaluation metrics (Table~\ref{tab:rank_corr_all}). 
Since higher noise levels degrade topic quality, a strong negative correlation is expected. 
AMI and ARI exhibited correlation coefficients close to $-1$, with $\rho \le -0.97$ across all four datasets. 
By contrast, the topic coherence metrics ($C_{\rm UCI}$, $C_{\rm NPMI}$, and $C_v$) showed correlations close to zero and even positive correlations for some datasets. For the 20News dataset, the coherence-based metrics also yielded correlation coefficients close to $-1$. 
These findings indicate that the coherence metrics may be unreliable for evaluating topic quality in short or informal texts, such as those found in social media.

\begin{table}[t]
  \centering
  \caption{{Spearman correlation coefficient between the noise level $p_n$ and the evaluation metrics for topic discovery. 
  Results are reported as mean $\pm$ standard deviation over five runs. 
  The mean values smaller than $-0.95$ is shown in {\bf bold}.}
  }
  \label{tab:rank_corr_all}
  \vspace{0.3cm}
\begin{tabular}{@{}lcccc@{}}
\toprule
\textbf{Metric} & \textbf{20News} & \textbf{AgNewsTitle} & \textbf{Reddit} & \textbf{TweetTopic} \\
\midrule
$C_v$ & \textbf{-0.992 $\pm$ 0.001} & 0.453 $\pm$ 0.033 & -0.260 $\pm$ 0.021 & -0.905 $\pm$ 0.012 \\
$C_{\mathrm{NPMI}}$ & \textbf{-0.993 $\pm$ 0.001} & -0.312 $\pm$ 0.012 & -0.229 $\pm$ 0.023 & -0.371 $\pm$ 0.060 \\
$C_{\mathrm{UCI}}$ & \textbf{-0.991 $\pm$ 0.002} & 0.073 $\pm$ 0.052 & -0.237 $\pm$ 0.019 & 0.075 $\pm$ 0.048 \\
AMI & \textbf{-0.995 $\pm$ 0.000} & \textbf{-0.978 $\pm$ 0.000} & \textbf{-0.995 $\pm$ 0.000} & \textbf{-0.995 $\pm$ 0.000} \\
ARI & \textbf{-0.995 $\pm$ 0.000} & \textbf{-0.986 $\pm$ 0.003} & \textbf{-0.995 $\pm$ 0.000} & \textbf{-0.995 $\pm$ 0.000} \\
\bottomrule
\end{tabular}
\end{table}

\clearpage

\section{ARI score to evaluate topic quality}
\label{appendix_ari}

\begin{figure}[t]
  \centering
  \includegraphics[width= 15cm]{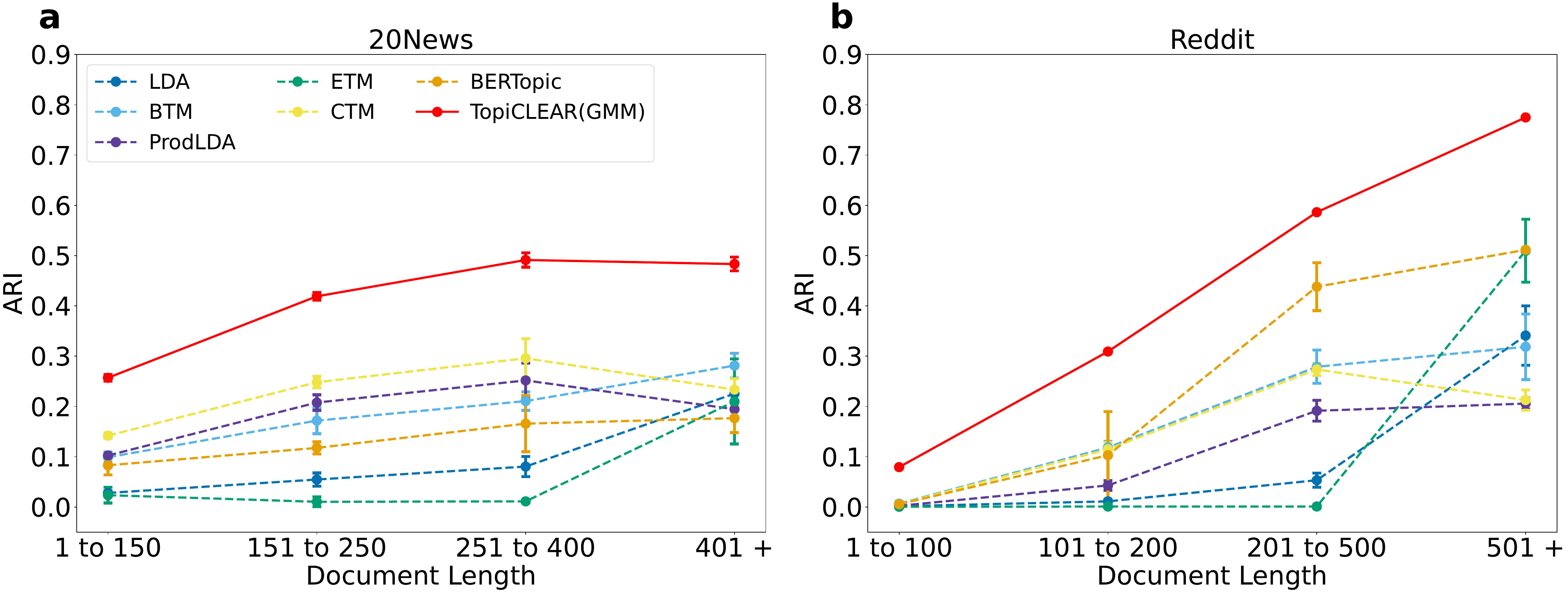}
  \vspace{0.2cm}
  \caption{
  Dependence of the ARI score on document length (word count). Two datasets are examined: (a) 20News and (b) Reddit. {Error bars indicate one standard deviation over five runs.} 
  }  
  \label{fig:len_ari}
\end{figure}

We calculate the ARI score to evaluate the topic quality identified by the proposed {flamework} (TopiCLEAR) as well as the five baseline methods and a reference method.  
Table~\ref{tab:score_ARI} shows the ARI scores for all methods across four datasets (20News, AgNewsTitle, Reddit, and TweetTopic). Consistent with the AMI results (Table~\ref{tab:score_AMI}), TopiCLEAR achieves the highest ARI on all four datasets. 
%   HDBSCAN 
{
The ARI values for BERTopic are substantially lower than those of the other baseline methods on the 20News, AgNewsTitle, and TweetTopic datasets.  
We therefore analyzed the topic granularity produced by BERTopic (HDBSCAN) (Table~\ref{tab:topic_numbers_BERTopic}).
The results reveal substantial discrepancies between the topic granularities defined by human annotations and those inferred by BERTopic. 
For example, BERTopic identified 3,561 topics on the AgNewsTitle dataset, compared with only four human-annotated topics. 
Moreover, more than half of the documents were assigned to the noise class in the AgNewsTitle, Reddit, and TweetTopic datasets. 
These findings highlight the difficulty of estimating human-interpretable topic granularity from document embeddings. 
} 
%   Fig. 4:  Document length dependency
Furthermore, we examined how document length influences ARI performance (Figure~\ref{fig:len_ari}). Similar to the AMI results (Fig.~\ref{fig:len_ami}), TopiCLEAR achieves the highest score for all document-length ranges.

\begin{table}[h]
\centering
\begin{threeparttable}
\caption{
{Topic discovery quality quantified by Adjusted Rand Index (ARI). 
Results are reported as mean $\pm$ standard deviation over five runs. The highest mean value is shown in {\bf bold}.}
}
\label{tab:score_ARI}
\vspace{0.3cm}
\begingroup
\setlength{\tabcolsep}{3.2pt}
\small
\begin{tabular}{@{}lcccc@{}}
\toprule
\textbf{Method} & \textbf{20News} & \textbf{AgNewsTitle} & \textbf{Reddit} & \textbf{TweetTopic} \\
\midrule
LDA & 0.212 $\pm$ 0.029 & 0.042 $\pm$ 0.012 & 0.215 $\pm$ 0.029 & 0.025 $\pm$ 0.014 \\
BTM & 0.250 $\pm$ 0.024 & 0.373 $\pm$ 0.062 & 0.198 $\pm$ 0.033 & 0.147 $\pm$ 0.012 \\
ProdLDA & 0.161 $\pm$ 0.014 & 0.130 $\pm$ 0.022 & 0.073 $\pm$ 0.011 & 0.125 $\pm$ 0.009 \\
ETM & 0.257 $\pm$ 0.041 & 0.026 $\pm$ 0.017 & 0.177 $\pm$ 0.101 & 0.244 $\pm$ 0.021 \\
CTM & 0.184 $\pm$ 0.010 & 0.211 $\pm$ 0.016 & 0.105 $\pm$ 0.008 & 0.181 $\pm$ 0.007 \\
BERTopic\tnote{a} & 0.028 $\pm$ 0.006 & 0.006 $\pm$ 0.001 & 0.154 $\pm$ 0.002 & 0.023 $\pm$ 0.005 \\
TopiCLEAR(GMM) & \textbf{0.401 $\pm$ 0.011} & \textbf{0.533 $\pm$ 0.000} & \textbf{0.363 $\pm$ 0.000} & \textbf{0.264 $\pm$ 0.004} \\
\bottomrule
\end{tabular}
\endgroup
\begin{tablenotes}[flushleft]
\footnotesize
\item[a] BERTopic is not directly comparable to the other methods, because it does not allow us to specify the number of topics beforehand.
\end{tablenotes}
\end{threeparttable}
\end{table}

\begin{table}[h]
\centering
\caption{
{Topic granularity obtained by BERTopic (via HDBSCAN). The number of human-annotated topics, the number of topics identified by BERTopic (excluding noise labels), and the proportion of documents assigned to the noise class are reported. Results are reported as mean $\pm$ standard deviation over five runs. } 
}
%%  Granularity of the HDBSCAN solutions in the clustering ablation. The number of clusters excludes the noise label.}
\label{tab:topic_numbers_BERTopic}
\vspace{0.3cm}
\begingroup
\setlength{\tabcolsep}{5pt}
\scriptsize
\begin{tabular}{@{}lccc@{}}
\toprule
\textbf{Dataset} & \textbf{\# Human-annotated topics} & \textbf{\# BERTopic topics} & \textbf{Noise rate} \\
\midrule
20News & 20 & 323 $\pm$ 12  & 37.6\% $\pm$ 0.8\% \\
AgNewsTitle & 4 & 1991 $\pm$ 36 & 34.0\% $\pm$ 0.4\% \\
Reddit & 5 & 291 $\pm$ 4   & 45.4\% $\pm$ 0.3\% \\
TweetTopic & 6 & 103 $\pm$ 3   & 41.8\% $\pm$ 1.6\% \\
%%  20News & 20 & 322.8 $\pm$ 11.9  & 37.6\% $\pm$ 0.8\% \\
%%  AgNewsTitle & 4 & 1990.6 $\pm$ 36.0 & 34.0\% $\pm$ 0.4\% \\
%%  Reddit & 5 & 291.2 $\pm$ 4.4   & 45.4\% $\pm$ 0.3\% \\
%%  TweetTopic & 6 & 103.2 $\pm$ 2.7   & 41.8\% $\pm$ 1.6\% \\
\bottomrule
\end{tabular}
\endgroup
\end{table}

\clearpage

\section{Sentence embeddings used in the ablation study}
\label{appendix_embedding}

In Ablation study (Sec.~\ref{section_ablation}), we used \texttt{paraphrase‑MiniLM‑L6‑v2}~\footnote{https://huggingface.co/sentence-transformers/paraphrase-MiniLM-L6-v2}, and  \texttt{intfloat/e5-small-v2}~\footnote{https://huggingface.co/intfloat/e5-small-v2} in addition to \texttt{all-MiniLM-L6-v2}. 
Both \texttt{all-MiniLM-L6-v2} and \texttt{paraphrase-MiniLM-L6-v2} are based on the same MiniLM-L6 architecture~\cite{Wang_2020}, which is a 6-layer Transformer encoder with 384-dimensional representations; however, they differ in the training objectives and the types of training data used. 
The \texttt{all-MiniLM-L6-v2} model is trained as a general-purpose sentence encoder using a large-scale mixture of sentence-pair datasets. In contrast, the \texttt{paraphrase-MiniLM-L6-v2} model is trained on datasets related to paraphrase identification and semantic similarity. 
The \texttt{intfloat/e5-small-v2} model is also a general-purpose text embedding model, based on a 12-layer Transformer encoder and producing 384-dimensional representations~\cite{wang2022text}.
\clearpage

\section{Ablation study using Adjusted Rand Index (ARI) }
\label{appendix_ablation} 

In addition to the AMI score, we conduct an ablation study of TopiCLEAR using the ARI score. 
%  Embedding 
First, we evaluate the impact of the sentence embedding on ARI (Table~\ref{tab:ablation_emb_ari}). 
As with the AMI results, {
\texttt{all-MiniLM-L6-v2} achieved the highest ARI on AgNewsTitle and Reddit datasets. 
\texttt{all-MiniLM-L6-v2} achieved similar ARI to \texttt{intfloat/e5-small-v2} on 20News and TweetTopic datasets.}  
%  Number of topics
Second, we examine the effect of the number of topics $K$ on ARI (Table~\ref{tab:ablation_k_ari}). Except for TweetTopic dataset, ARI also takes the highest value when $K$ matches its true value. 
%% We find that the number of topics at which $C_v$ is maximized provides an estimate of the approximate number of topics. However, this estimate slightly overestimates the true number of topics. 

%  ADR
{Next, we examine the effect of Adaptive Dimensionality Reduction (ADR) by comparing the ARI score at the initial state with that after convergence (Table~\ref{tab:ablation_clust_ari}).  
While ADR consistently improves the ARI score for two formal-text datasets (20News and AGNews),  
its effect is not significant for informal-text datasets (Reddit and TweetTopic).}  
%%  Furthermore, we examine how  the choice of the clustering method affects ARI (Table~\ref{tab:ablation_clust_ari}). Consistent with the AMI retult (Table~\ref{tab:ablation_clust_ami}), methods with a fixed number of clusters---$k$‑means, GMM, and their ADR-enhanced versions---achieve substantially higher ARI scores than HDBSCAN. While ADR also improves ARI performance, the most suitable clustering method ($k$-means or GMM) varies across datasets. 
%  Preprocessing 
Table~\ref{tab:ablation_pre_ari} reports the effect of PCA-based preprocessing on ARI. 
PCA-based preprocessing generally improves ARI, although the effect is dataset-dependent. Although the PCA dimension $D$ that maximizes ARI differs across datasets, ARI is only weakly dependent on $D$.
\vspace{0.5cm}

%  Ablation study:  Sentence Embeddings
\begin{table}[h]
\centering
\caption{
{Effect of sentence embedding on ARI. 
Results are reported as mean $\pm$ standard deviation over five runs. The highest mean value is shown in {\bf bold}. }
}
%%{Effect of sentence embedding models on the ARI score. Values are reported as the mean $\pm$ sample standard deviation over five runs. The largest unrounded mean is highlighted in {\bf bold}.}
\label{tab:ablation_emb_ari}
\begingroup
\setlength{\tabcolsep}{2.5pt}
\footnotesize
\begin{tabular}{@{}lcccc@{}}
\toprule
\textbf{Embedding} & \textbf{20News} & \textbf{AgNewsTitle} & \textbf{Reddit} & \textbf{TweetTopic} \\
\midrule
\texttt{all-MiniLM-L6-v2} & \textbf{0.401 $\pm$ 0.011} & \textbf{0.533 $\pm$ 0.000} & \textbf{0.363 $\pm$ 0.000} & \textbf{0.264 $\pm$ 0.004} \\
\texttt{intfloat/e5-small-v2} & 0.399 $\pm$ 0.026 & 0.417 $\pm$ 0.000 & 0.264 $\pm$ 0.000 & \textbf{0.264 $\pm$ 0.010} \\
\texttt{paraphrase-MiniLM-L6-v2} & 0.342 $\pm$ 0.013 & 0.439 $\pm$ 0.000 & 0.300 $\pm$ 0.000 & 0.228 $\pm$ 0.035 \\
\bottomrule
\end{tabular}
\endgroup
\end{table}

%  Ablation study:  the number of topics K
\begin{table}[t]
\centering
\caption{
{Effect of the number of topics $K$ on ARI and \textbf{$C_v$}. 
Results are reported as mean $\pm$ standard deviation over five runs. The highest mean value is shown in {\bf bold}. The true number of topics is highlighted in \magenta{magenta}.} 
}
%% {Effect of the number of topics $K$ on ARI and \textbf{$C_v$}. Values are reported as the mean $\pm$ sample standard deviation over five runs. The largest unrounded mean within each dataset and metric is highlighted in {\bf bold}. The true number of topics is highlighted in \magenta{magenta}. }}
\label{tab:ablation_k_ari}
\vspace{0.3cm}
\begingroup
\setlength{\tabcolsep}{5pt}
\begin{tabular}{@{}llcc@{}}
\toprule
\textbf{Dataset} & \textbf{$K$} & \textbf{ARI} & \textbf{$C_v$} \\
\midrule
20News & 5 & 0.200 $\pm$ 0.000 & 0.514 $\pm$ 0.000 \\
 & 10 & 0.365 $\pm$ 0.007 & 0.524 $\pm$ 0.004 \\
 & \magenta{20} & \textbf{0.401 $\pm$ 0.011} & 0.552 $\pm$ 0.007 \\
 & 30 & 0.376 $\pm$ 0.007 & 0.568 $\pm$ 0.017 \\
 & 40 & 0.339 $\pm$ 0.008 & \textbf{0.584 $\pm$ 0.002} \\
\midrule
AgNewsTitle & 2 & 0.260 $\pm$ 0.000 & 0.330 $\pm$ 0.000 \\
 & \magenta{4} & \textbf{0.533 $\pm$ 0.000} & 0.351 $\pm$ 0.000 \\
 & 6 & 0.346 $\pm$ 0.000 & \textbf{0.389 $\pm$ 0.004} \\
 & 8 & 0.331 $\pm$ 0.006 & 0.355 $\pm$ 0.004 \\
 & 10 & 0.265 $\pm$ 0.001 & 0.379 $\pm$ 0.017 \\
\midrule
Reddit & 3 & 0.293 $\pm$ 0.000 & 0.437 $\pm$ 0.000 \\
 & \magenta{5} & \textbf{0.363 $\pm$ 0.000} & 0.433 $\pm$ 0.000 \\
 & 8 & 0.351 $\pm$ 0.003 & 0.434 $\pm$ 0.003 \\
 & 10 & 0.303 $\pm$ 0.000 & 0.449 $\pm$ 0.003 \\
 & 13 & 0.254 $\pm$ 0.009 & \textbf{0.452 $\pm$ 0.003} \\
\midrule
TweetTopic & 3 & \textbf{0.499 $\pm$ 0.002} & 0.434 $\pm$ 0.000 \\
 & \magenta{6} & 0.264 $\pm$ 0.004 & 0.411 $\pm$ 0.004 \\
 & 9 & 0.243 $\pm$ 0.018 & \textbf{0.438 $\pm$ 0.008} \\
 & 12 & 0.191 $\pm$ 0.026 & 0.434 $\pm$ 0.019 \\
 & 15 & 0.167 $\pm$ 0.004 & 0.422 $\pm$ 0.009 \\
\bottomrule
\end{tabular}
\endgroup
\end{table}
\clearpage 

%  Ablation study:  Clustering
\begin{table}[h]
\centering
\caption{ 
{
Effect of ADR on ARI. Results are reported as mean $\pm$ standard deviation over five runs. The highest mean value is shown in {\bf bold}. } 
}
%%{Effect of clustering methods on the ARI score. Values are reported as the mean $\pm$ sample standard deviation over five runs. The largest unrounded mean is highlighted in {\bf bold}.}}
\label{tab:ablation_clust_ari}
\begingroup
\setlength{\tabcolsep}{2.5pt}
\small
\begin{tabular}{@{}lcccc@{}}
\toprule
 & \textbf{20News} & \textbf{AgNewsTitle} & \textbf{Reddit} & \textbf{TweetTopic} \\
\midrule
   & \multicolumn{4}{c}{\bf $k$-means} \\
   Initial  & 0.412 $\pm$ 0.004 & 0.485 $\pm$ 0.000 & 0.359 $\pm$ 0.000 & 0.251 $\pm$ 0.002 \\
   Converged & {\bf 0.457 $\pm$ 0.009} & 0.512 $\pm$ 0.000 & 0.349 $\pm$ 0.000 & {\bf 0.289 $\pm$ 0.001} \\
   & \multicolumn{4}{c}{\bf GMM} \\
   Initial  & 0.390 $\pm$ 0.008 & 0.485 $\pm$ 0.000 & {\bf 0.370 $\pm$ 0.000} & 0.259 $\pm$ 0.022 \\
   Converged  & 0.401 $\pm$ 0.011 & {\bf 0.533 $\pm$ 0.000} & 0.363 $\pm$ 0.000 & 0.264 $\pm$ 0.004 \\
%%HDBSCAN & 0.000 $\pm$ 0.000 & 0.000 $\pm$ 0.000 & 0.025 $\pm$ 0.000 & 0.100 $\pm$ 0.000 \\
%%$k$-means & 0.417 $\pm$ 0.032 & 0.483 $\pm$ 0.045 & 0.377 $\pm$ 0.014 & 0.266 $\pm$ 0.016 \\
%%GMM & 0.440 $\pm$ 0.013 & 0.525 $\pm$ 0.000 & \textbf{0.388 $\pm$ 0.004} & 0.286 $\pm$ 0.016 \\
%%ADR: $k$-means & \textbf{0.458 $\pm$ 0.010} & 0.512 $\pm$ 0.000 & 0.349 $\pm$ 0.000 & \textbf{0.289 $\pm$ 0.000} \\
%%ADR: GMM & 0.401 $\pm$ 0.011 & \textbf{0.533 $\pm$ 0.000} & 0.363 $\pm$ 0.000 & 0.264 $\pm$ 0.004 \\
\bottomrule
\end{tabular}
\endgroup
\end{table}

\begin{table}[h]
\centering
\caption{
{Effect of preprocessing step on ARI. 
Results are reported as mean $\pm$ standard deviation over five runs. 
{\bf Bold} and \underline{underlined} values indicate the highest and second-highest mean values, respectively.} 
}
%%  {Effect of preprocessing step on the ARI score. Values are reported as the mean $\pm$ sample standard deviation over five runs. The largest and second-largest unrounded means are highlighted in {\bf bold} and the second best is \underline{underlined}.}}
\label{tab:ablation_pre_ari}
\begingroup
\setlength{\tabcolsep}{3.2pt}
\small
\begin{tabular}{@{}lcccc@{}}
\toprule
\textbf{Preprocessing} & \textbf{20News} & \textbf{AgNewsTitle} & \textbf{Reddit} & \textbf{TweetTopic} \\
\midrule
\textbf{PCA: $D{=}32$} & 0.399 $\pm$ 0.018 & \underline{0.533 $\pm$ 0.000} & 0.353 $\pm$ 0.000 & \underline{0.268 $\pm$ 0.014} \\
\textbf{PCA: $D{=}64$} & \underline{0.401 $\pm$ 0.011} & \textbf{0.533 $\pm$ 0.000} & \underline{0.363 $\pm$ 0.000} & 0.264 $\pm$ 0.004 \\
\textbf{PCA: $D{=}128$} & \textbf{0.422 $\pm$ 0.015} & 0.509 $\pm$ 0.000 & \textbf{0.364 $\pm$ 0.000} & 0.267 $\pm$ 0.008 \\
\textbf{None} & 0.399 $\pm$ 0.006 & 0.472 $\pm$ 0.000 & 0.360 $\pm$ 0.000 & \textbf{0.275 $\pm$ 0.001} \\
\bottomrule
\end{tabular}
\endgroup
\end{table}
\clearpage

\section{Representative tweets for topics in the “sports \& gaming” subset of the TweetTopic dataset}
\label{appendix_output}

In this appendix, we present the results of applying TopiCLEAR(GMM) and LDA to the “sports \& gaming” subset of the TweetTopic dataset. 
Tables~\ref{tab:representative_tweets_topiclear} and ~\ref{tab:representative_tweets_lda} show five representative tweets on each topic obtained by TopiCLEAR and LDA, respectively. 
For each topic, five tweets with the highest assignment probabilities are shown.
\vspace{1cm}

\scriptsize
\setlength{\tabcolsep}{1pt}
\renewcommand{\arraystretch}{1}
\setlength{\LTleft}{-0.05\linewidth}
\setlength{\LTright}{-0.05\linewidth}

\begin{longtable}{@{}>{\centering\arraybackslash}p{0.1\linewidth}>{\raggedright\arraybackslash}p{1\linewidth}@{}}
\caption{
Representative tweets assigned to each TopiCLEAR(GMM) topic in the ``sports \& gaming'' subset of the TweetTopic dataset. 
}

\label{tab:representative_tweets_topiclear}\\
\toprule
Topic & Representative tweet \\
\midrule
\endfirsthead

\toprule
Topic & Representative tweet \\
\midrule
\endhead

\endfoot

\bottomrule
\endlastfoot
\centering\textbf{A}\par{\scriptsize $n=1,052$} & \textbullet\ ICYMI: Melvin Ingram already thinks this could be a special year for the Steelers.   “The sky is the limit, not only for this defense, but for this team.”  From: \{@Ray Fittipaldo@\} \{\{URL\}\}  \\
 & \textbullet\ So Doug Williams and Co. didn’t score 28 in a quarter in the Super Bowl. Jan 26, 1992! What I miss? \{@Sunday Night Football on NBC@\}  \\
 & \textbullet\ Our national writers made their cases for who should be the fourth team in the College Football Playoff, after Alabama, Clemson and Ohio State.   Yo, \#Sooners fans, go follow \{@Ari Wasserman@\} . He s your friend. \{\{URL\}\}  \\
 & \textbullet\ The Texans were up 24-0 headed into the second quarter. Going into the locker room, the Chiefs are leading 28-24. Most teams can’t make up that difference over the course of an entire game. WHAT A HALF! \#NFL | \#NFLPlayoffs | \#HOUvsKC | \#WeAreTexans | \#ChiefsKingdom \{@NFL on CBS@\}  \\
 & \textbullet\ Blah blah blah, tired reasons and excuses.  I am just a bitter Niners fan, so I m not rooting for the Bucs, I m just rooting against the Chiefs.  Hope it ends well for you primo, \{\{USERNAME\}\} . \\
\addlinespace[1.5em]
\centering\textbf{B}\par{\scriptsize $n=514$} & \textbullet\ “These facilities have the stature of a Football League club.”  We sat down with \{@Wales@\} Euro 2016 semi-finalist \{@Dave Edwards@\} following his visit to the club yesterday afternoon \\
 & \textbullet\ Dreadful firs t half from \{@Tottenham Hotspur@\} . Ball was in the air way too much with no real control. Need to find some midfield without our ‘best’ one who is already injured.  Not the start to the decade we are looking for let’s see if \{@Harry Kane@\} and company can turn it around!! \#COYS \\
 & \textbullet\ Congratulations To Trezeguet and Mohamedy and Team Aston Villa Stay in Premier League \{@Trezeguet@\} \{@Aston Villa@\} \\
 & \textbullet\ Hahahahaha can’t even win the bottom 2 Premier League teams. What a joke \{@Manchester United@\} \\
 & \textbullet\ I struggle to understand why \{@Manchester United@\} fans are surprised about results. The manager managed Cardiff which he was very poor and Molde, even if he was good that league can’t be compared to this Eng league. And you all thought he would be better than Mourinho and Van Gaal...\#deluded \\
\addlinespace[1.5em]
\centering\textbf{C}\par{\scriptsize $n=261$}  & \textbullet\ UFC 249 was a success so will boxing be back soon? \{\{URL\}\} via \{@WordPress com@\} \#boxing \#UFC249 \#lockdown \\
 & \textbullet\ Check out the Fight Breakdown of \{\{USERNAME\}\} and \{@Robert Whittaker@\} \#UFC254 \{\{URL\}\} \\
 & \textbullet\ Props to Alex but \{@Max Holloway@\} won that fight \#UFC251 \#blessedexpress \#realtalk \\
 & \textbullet\ Column: Two years ago, \{@TYSON FURY@\} was lost. After knocking out Deontay Wilder, he is heavyweight champion, again. On a boxing redemption story \{\{URL\}\}  \\
 & \textbullet\ UFC 265: Francis Ngannou, Jon Jones, Conor McGregor react to Ciryl Gane s interim championship win - via \{@ESPN@\} App \{\{URL\}\} \\ \midrule
 \addlinespace[-0.1em]
\centering\textbf{D}\par{\scriptsize $n=656$} &  \textbullet\ Finally \{\{USERNAME\}\} \{\{USERNAME\}\} I finished ff14 what an amazing ride from AAR to Shadowbringers the story’s and characters are some of the best I experienced in gaming, and I’m glad y’all got me back into it thank you . \\
 & \textbullet\ So yesterday I installed Genshin Impact because \{\{USERNAME\}\} wouldn t shut up and now I love her for not shutting up thank you for showing me this awesome game \#GenshinImpact \\
 & \textbullet\ Had a great day of doing demos of Tiny Epic Tactics for \{\{USERNAME\}\} and \{\{USERNAME\}\} ! A lot of happy people, a lot of people picked up the game (sold out all but one copy in the store!) It s really awesome to connect people with a new game. :) \\
 & \textbullet\ Omg. That was the greatest \{@The CrossFit Games@\} ever. Never have I been so emotional about it. Thank you \{@CrossFit@\} \{\{USERNAME\}\} \\
 & \textbullet\ I, along with many other \#PokkenTournamentDX members, are answering questions related to the game asked on \{\{USERNAME\}\} !   If you re new and want to know more about anything Pokkén, I highly recommend shooting some questions! \{\{URL\}\}  \\
\end{longtable}
\clearpage

\begin{longtable}{@{}>{\centering\arraybackslash}p{0.1\linewidth}>{\raggedright\arraybackslash}p{0.98\linewidth}@{}}
\caption{
Representative tweets assigned to each LDA topic in the ``sports \& gaming'' subset of the TweetTopic dataset. 
} 
\label{tab:representative_tweets_lda}\\
\toprule
Topic  &  Representative tweet \\
\midrule
\endfirsthead

\toprule
Topic  &  Representative tweet \\
\midrule
\endhead

\endfoot

\bottomrule
\endlastfoot
\centering\textbf{A}\par{\scriptsize $n=648$} &  \textbullet\ Rohit As An Opener  * IN ODI - Played 1st Match Vs SA (2011) * In Test - Going To Play First Match As Opener Vs SA (2019)* * T20 Debut IN SA * Maiden T20 50 Vs SA * Maiden ODI 50 As An Opener Vs SA  * Maiden T20I 100 Vs SA All D Best Champ \{@Rohit Sharma@\} \#IndvsSA \\
  & \textbullet\ PREVIEW: Even though the first round is still going on, it’s already time to get into the division finals as the New York Islanders ( \{@New York Islanders@\} ) \& Boston \{@Boston Bruins@\} battle for a spot in the final four.   Here’s how I see the series playing out. \{\{URL\}\}  \\
  & \textbullet\ Kamaru Usman Vs Colby Covington Full Fight UFC 245 Highlights 2019 \{\{URL\}\} via \{@YouTube@\} . Every entity or persons associated with Donald wld be resoundingly defeated @ the polls \& in every endeavour of life. According 2 Rick Wilson every thing trump touches wld die \\
  & \textbullet\  \{@Kurt Warner@\} : With the Greatest Show on Turf \{@Los Angeles Rams@\} , my WR getting was about SEPARATION. With \{@Arizona Cardinals@\} , my WR getting open was about POSITIONING. \#TightWindow \#FootballOpen \#NFLopen   always great, gotta love it ! \{@NFL Network@\} \{@Las Vegas Raiders@\}  \{@Miami Dolphins@\}  \\
  & \textbullet\ NFL officials love the replay but why can’t they get it right the first time. Ball clearly out before crossing the line and the \{@NFL@\} \{@NFL Officiating@\} rule a touchdown. If they don’t say fumble and clear recovery by Miami, this will be another example of terrible officiating \\
\addlinespace[1.5em]
\centering\textbf{B}\par{\scriptsize $n=467$}  & \textbullet\ Things I know: I like Freddie Kitchens. Freddie is a new head coach who hasn t found his way yet. He has a young QB who is struggling. The defense is better. \{@Cleveland Browns@\} deserve to be 1-2. My prediction of 7-8-1 this season is looking good. We can win the Super Bowl soon. Not 2019. \\
  & \textbullet\ I’m told Chelsea will debut their new home kit in the FA Cup final vs Leicester at Wembley on Saturday. No decision yet as to whether they’ll wear it in the Champions League final  vs Manchester City later this month. \{@Liam Twomey@\} \\
  & \textbullet\ 10 points from 21 including games against the bottom 3, not good enough. The lack of desire and fight from some players is telling,  too many going missing during games. The onus is on the forwards to get goals and the forwards at \{@Manchester United@\} are failing big time. \#MUFC \\
  & \textbullet\ Join \{@Mark Zinno@\} and myself as we broadcast live from Life University this afternoon from noon to three. Few things we’ll get to...  Did Zinno make money off the XFL yesterday? Kevin Durant is still soft. Zinno on the Red Carpet National Signing Day \{@680 The Fan@\} \\
  & \textbullet\ I just want \{@Los Angeles Gladiators@\} fans and team to know that you put up such a strong fight for a slot to Hawaii. You pushed Fuel to make small changes with your comp changes. You were competitive and made the series interesting and fun to watch.  GGs and hoping to see more of you guys! \\
\addlinespace[1.5em]
\centering\textbf{C}\par{\scriptsize $n=600$} & \textbullet\ Packers vs Cowboys Live Stream - <NFL Football> \{@Free@\} tv link :>> \{\{URL\}\} Green Bay Packers vs. Dallas Cowboys live-stream info, TV channel.Cowboys vs Packers Sunday s NFL football game. Cowboys vs. Packers [live] <> stream. \\
  & \textbullet\ Cowboys vs. Giants LIVE Week 17 watch party, presented by \{@MANSCAPEDTM@\} A win keeps the Cowboys  playoff chances alive until at least Sunday Night Football  Come join us on \{\{USERNAME\}\}   Dallas Cowboys Report: \{\{URL\}\}  \\
  & \textbullet\ Eagles coach Doug Pederson left it simple that Jalen Hurts is the starter for this game vs. \{@New Orleans Saints@\} but player and team sources believe it is for rest of season, barring injury or disastrous play. But, as one player noted, it s been a disaster anyway with too many issues to count \\
  & \textbullet\ MMA vs \{@Cowboy Cerrone@\} Link    \{\{URL\}\} UFC 246 McGregor vs Cowboy Live Stream Free  Date: Saturday, January 18, 2020  \#McGregorCerrone \#UFC \#CerroneMcGregor \#mma \#boxing \#live \#stream \#free \#online \#Fighting \\
  & \textbullet\ Michael Chandler vs Charles Oliveira Live Stream  Charles Oliveira vs Michael Chandler free stream online UFC 262 Lightweight Some minutes to go  . . . .   WATCH LIVE via \{\{USERNAME\}\} WATCH LIVE via \{\{USERNAME\}\} \\
\midrule
 \addlinespace[-0.1em]
\centering\textbf{D}\par{\scriptsize $n=768$}  & \textbullet\ Preparing Raymond James Stadium Super Bowl LV \{\{URL\}\} via \{@10 Tampa Bay@\} They need to cancel the Super bowl here in Florida. That s crazy to hold in Florida where the virus is  is spreading faster every day !! There is a bad bad pandemic case load here in Florida. \\
  & \textbullet\ Said all season the \#Bills are a year away. Playoffs this year was a bonus \& a good one. It’s a young, core group, led by the QB. Give Allen another year to develop, keep the defense in-tact, draft a WR, use cap space wisely \& the \{@Buffalo Bills@\} are set up for long-term success. \\
  & \textbullet\ Everyone picking the \#Titans today to cover seem to ignore the \{@Baltimore Ravens@\} scored 40+ five times, haven’t lost since September \& are 14-2 with an MVP QB. The Titans beat up a collapsing \#Patriots team last week, but let’s not overplay it. Bet the \#Ravens! \#NFLPlayoffs \#NFL \#BALvsTEN \\
  & \textbullet\ 10 days ago the seahawks squeaked out a win against the rams. 6 days ago the \{@San Francisco 49ers@\} dominated the Browns. Today the seahawks barely won against the same Browns and the Niners erased that rams offense with less time to prepare. What excuses will the national media find this week? \\
  & \textbullet\ Everyone knows how big of a \{@Tom Brady@\} fan I am. SUPERFAN. So you know I’m excited about this game. Everyone was saying I was crazy for saying the \{@Tampa Bay Buccaneers@\} would make the Super Bowl this year. People need to quit trying to debate me. I know my stuff obviously. \\
\end{longtable}

\end{appendices}

\end{document}